\documentclass[sigconf,nonacm,pdfa]{acmart}

\usepackage{amsmath,amsfonts,amsthm}
\usepackage{graphicx}
\usepackage{textcomp}
\usepackage{balance}
\usepackage{color,xcolor}
\usepackage{multirow}
\usepackage{subfigure}
\usepackage{array}
\usepackage{enumitem}
\usepackage{epstopdf}
\usepackage{booktabs}
\usepackage{threeparttable}
\usepackage{bm}
\usepackage{hyperref}
\usepackage{xr}
\usepackage[a-2b,mathxmp]{pdfx}[2018/12/22] 

\newtheoremstyle{mydefn}
{}{}
{\it}       
{0pt}       
{\bfseries} 
{:~}        
{0pt}       
{}          
\theoremstyle{mydefn}

\newtheoremstyle{myexample}
{}{}
{}          
{0pt}       
{\bfseries} 
{:~}        
{0pt}       
{}          
\theoremstyle{myexample}

\newtheorem{example}{Example}

\renewcommand{\paragraph}[1]{\vspace{0.5em}\noindent\textbf{#1.}}
\renewcommand{\subparagraph}[1]{\vspace{0.25em}\noindent\textit{\underline{#1.}}}

\newcommand{\norm}[1]{\Vert #1 \Vert}
\newcommand{\num}[1]{\vert #1 \vert}

\makeatletter
\newif\if@restonecol
\makeatother

\usepackage[linesnumbered,ruled,vlined]{algorithm2e}
\usepackage{algpseudocode}

\SetKwComment{Comment}{$\triangleright$\ }{}
\SetKwRepeat{Do}{do}{while}

\newcommand\vldbdoi{XX.XX/XXX.XX} 
\newcommand\vldbpages{XXX-XXX}    
\newcommand\vldbvolume{17} 
\newcommand\vldbissue{8}   
\newcommand\vldbyear{2024} 
\newcommand\vldbauthors{\authors}
\newcommand\vldbtitle{\shorttitle} 
\newcommand\vldbavailabilityurl{https://github.com/e0001428/local-distribution-shifting-synthesis/}
\newcommand\vldbpagestyle{empty} 

\begin{document}

\title{From Zero to Hero: Detecting Leaked Data through Synthetic Data Injection and Model Querying}

\author{Biao Wu}
\orcid{0000-0003-4240-4647}
\affiliation{%
  \institution{National University of Singapore}
  \country{} 
}
\email{wubiao@comp.nus.edu.sg}

\author{Qiang Huang}
\authornote{Qiang Huang is the corresponding author.}
\orcid{0000-0003-1120-4685}
\affiliation{%
  \institution{National University of Singapore}
  \country{} 
}
\email{huangq@comp.nus.edu.sg}

\author{Anthony K. H. Tung}
\orcid{0000-0001-7300-6196}
\affiliation{%
  \institution{National University of Singapore}
  \country{} 
}
\email{atung@comp.nus.edu.sg}

\begin{abstract}
Safeguarding the Intellectual Property (IP) of data has become critically important as machine learning applications continue to proliferate, and their success heavily relies on the quality of training data. While various mechanisms exist to secure data during storage, transmission, and consumption, fewer studies have been developed to detect whether they are already leaked for model training without authorization. This issue is particularly challenging due to the absence of information and control over the training process conducted by potential attackers. 

In this paper, we concentrate on the domain of tabular data and introduce a novel methodology, Local Distribution Shifting Synthesis (\textsc{LDSS}), to detect leaked data that are used to train classification models. The core concept behind \textsc{LDSS} involves injecting a small volume of synthetic data--characterized by local shifts in class distribution--into the owner's dataset. This enables the effective identification of models trained on leaked data through model querying alone, as the synthetic data injection results in a pronounced disparity in the predictions of models trained on leaked and modified datasets. \textsc{LDSS} is \emph{model-oblivious} and hence compatible with a diverse range of classification models. We have conducted extensive experiments on seven types of classification models across five real-world datasets. The comprehensive results affirm the reliability, robustness, fidelity, security, and efficiency of \textsc{LDSS}. Extending \textsc{LDSS} to regression tasks further highlights its versatility and efficacy compared with baseline methods.
\end{abstract}

\maketitle

\pagestyle{\vldbpagestyle}
\begingroup\small\noindent\raggedright\textbf{PVLDB Reference Format:}\\
\vldbauthors. \vldbtitle. PVLDB, \vldbvolume(\vldbissue): \vldbpages, \vldbyear.\\
\href{https://doi.org/\vldbdoi}{doi:\vldbdoi}
\endgroup
\begingroup
\renewcommand\thefootnote{}\footnote{\noindent
This work is licensed under the Creative Commons BY-NC-ND 4.0 International License. Visit \url{https://creativecommons.org/licenses/by-nc-nd/4.0/} to view a copy of this license. For any use beyond those covered by this license, obtain permission by emailing \href{mailto:info@vldb.org}{info@vldb.org}. Copyright is held by the owner/author(s). Publication rights licensed to the VLDB Endowment. \\
\raggedright Proceedings of the VLDB Endowment, Vol. \vldbvolume, No. \vldbissue\ %
ISSN 2150-8097. \\
\href{https://doi.org/\vldbdoi}{doi:\vldbdoi} \\
}\addtocounter{footnote}{-1}\endgroup

\ifdefempty{\vldbavailabilityurl}{}{
\vspace{.3cm}
\begingroup\small\noindent\raggedright\textbf{PVLDB Artifact Availability:}\\
The source code, data, and/or other artifacts have been made available at \url{\vldbavailabilityurl}.
\endgroup
}

\section{Introduction}
\label{sect:intro}

It is of paramount importance to protect data through strict enforcement of authorized access. 
Leaked data can have severe repercussions, ranging from exposure of private information to reputational harm.
More critically, it can lead to loss of competitive business advantages \cite{cheng2017enterprise}, violations of legal rights like privacy\cite{shklovski2014leakiness}, and breaches of national security \cite{bellia2011wikileaks}. 
Substantial efforts are directed towards bolstering data security, encompassing secure storage, safe data transfer, and watermarking to trace leaked data, thereby protecting the owner's Intellectual Property (IP) via legal channels.
Such watermarking techniques are typically employed in images and videos by embedding either visible or concealed information in the pixels to denote ownership \cite{huang2000embedding, potdar2005survey, cox2007digital, bi2007robust}. 

With the evolution of machine learning and artificial intelligence techniques, new and nuanced threats surface beyond conventional concerns about data theft and publication. 
Notably, malicious entities who gain access to leaked datasets often utilize them to train models, seeking a competitive edge. For instance, such training models can be exploited for commercial endeavors or to derive insights from datasets that were supposed to remain inaccessible. 
This emerging threat remains underexplored, and traditional digital watermarking methods might fall short in ensuring traceability, especially in a black-box setting where one can only access a model through a query interface \cite{ribeiro2015mlaas}.

In this paper, we attempt to address the emerging threat of detecting leaked data used for model training.
We introduce a novel method, Local Distribution Shifting Synthesis (\textsc{LDSS}), designed to achieve exemplary detection precision on such leakage.
Our study focuses on tabular datasets. 
For image and video datasets, existing work such as training models on watermarked or blended images \cite{zhang2018protecting} and label flipped images\cite{shafahi2018poison, adi2018turning} remain effective. This is because researchers often employ Deep Neural Networks (DNNs) on such datasets, where subtle watermarks can be seamlessly integrated without compromising model efficacy yet still be retained for identification purposes. 
Nonetheless, a substantial number of sensitive datasets, like those pertaining to patients or financial information, are stored in tabular format within relational databases. 
They typically possess fewer features compared to images with thousands of pixels. Without meticulous design, embedding a "watermark" into a limited number of columns could severely compromise model accuracy. 
Moreover, such alterations might go unnoticed by the machine learning model, especially when simpler models like Decision Tree, Support Vector Machine, and Random Forest are employed for these types of tabular datasets. 
Specifically, we concentrate on the classification task, as many functions in the financial and healthcare sectors hinge on classification. Examples include determining a patient's susceptibility to specific diseases based on screening metrics \cite{palaniappan2007biometrics, huang2007feature} as well as assessing if an applicant qualifies for a loan based on their data \cite{sheikh2020approach, bhatore2020machine}. 
Additionally, we assume that only black-box access to the target model is available. This is because unauthorized parties who engage in data theft typically do not disclose their model details, providing only a query interface instead.

Detecting leaked data utilized for model training solely through model querying is non-trivial. 
At first glance, employing the membership inference attack \cite{shokri2017membership} appears to be a straightforward yet effective solution. 
However, unauthorized entities can counteract this by applying model generalization techniques \cite{truex2021demystifying}. 
They might strategically prompt the model to produce incorrect predictions for queries that precisely match instances in their training data, thereby evading detection. 
Furthermore, this solution sometimes yields a high false-positive rate. This is especially true if the allegedly leaked data originates from a commonly accessed source. In such cases, the third party might coincidentally possess a similar dataset, which could have closely related or even identical instances.

There exist various watermarking and backdoor mechanisms \cite{li2021survey, xue2021intellectual, boenisch2021systematic} that are designed to embed information into target models. Subsequently, this information can be extracted with high precision to assert ownership. 
Nevertheless, most of these methods predominantly target the image classification task using DNN models \cite{zhang2020model, zhang2021deep, fei2022supervised, peng2023intellectual}. 
This preference arises from the rich parameters inherent in DNN models, which facilitate the embedding of additional information without compromising model performance. 
Their techniques, however, are not suitable for the problem of leaked data detection we explore in this paper. Tabular data, frequently used in fields like healthcare and finance where explainability is paramount, often employ non-DNN models. 
Moreover, dataset owners lack access to, control over, or even knowledge about the model training process in leaked data scenarios. 
Therefore, most extant methods in model watermarking \cite{darvish2019deepsigns, namba2019robust} and data poisoning or model backdoor \cite{chen2017targeted, saha2020hidden} fall short as they typically target a specific model type and necessitate access or control of the training process to guarantee optimal and effective embedding.

To overcome the above challenges, \textsc{LDSS} adopts a \emph{model-oblivious} mechanism, ensuring robust leaked data detection across different model types and training procedures.
The conceptual foundation of \textsc{LDSS} is inspired by the application of active learning and backdoor attacks used to embed watermarks into DNN models \cite{adi2018turning}. 
Rather than merely employing random label flipping, \textsc{LDSS} adopts a more nuanced synthetic data injection strategy. It injects synthetic samples into empty local spaces unoccupied by the original dataset, thereby altering local class distributions.
This technique obviates modifying or removing existing instances, preserving the original distribution within the occupied local spaces. 
Consequently, the classification model trained with these injected samples remains at a \emph{similar level of accuracy}; meanwhile, the synthetic injections produce predictions that are \emph{substantially different} in the affected local spaces compared to models trained without such injections. 
This ensures a low rate of false positives, even for models trained on original datasets or datasets derived from similar populations.

Once the owner modifies the dataset, any subsequent data acquired by an attacker will be this modified version.
The leaked data detection is executed by model querying, i.e., querying the suspect model using a trigger set of synthetic samples drawn from the same local space as the previously injected samples.
To rigorously assess the performance of \textsc{LDSS}, we employed a comprehensive set of evaluation metrics across seven types of classification models. 
While these models share common objectives, their underlying principles and distinctive characteristics offer a holistic perspective on \textsc{LDSS}'s performance under diverse conditions.
Our extensive experimental results, spanning five real-world datasets, showcase the superior performance of \textsc{LDSS} compared to two established baselines.
Importantly, these outcomes underscore its reliability, robustness, fidelity, security, and efficiency.
Additionally, we extend \textsf{LDSS} to regression tasks, conducting experiments on two real-world datasets and assessing its performance across seven regression models. This extension showcases \textsf{LDSS}'s versatility and leading efficacy across different tasks.

\paragraph{Organization}
The rest of this paper is organized as follows. 
The problem of leaked data detection is formulated in Section \ref{sect:problem}. We introduce the \textsc{LDSS} method in Section \ref{sect:method}. The experimental results and analyses are presented in Section \ref{sect:expt}. Section \ref{sect:related_work} reviews related work. Finally, we conclude this work in Section \ref{sect:conclusion}.

\section{Problem Formulation}
\label{sect:problem}

In this section, we begin by formally defining the problem of leaked data detection. Following that, we draw comparisons to the problem of model backdooring through data poisoning. Lastly, we outline the criteria for an ideal solution for leaked data detection. Table \ref{tab:notation} summarizes the frequently used notations in this paper.

\begin{table}[t]
\centering
\captionsetup{skip=0.5em}
\small
\caption{List of frequently used notations.}
\label{tab:notation}
\begin{tabular}{p{0.05\textwidth}p{0.38\textwidth}} \toprule 
  \textbf{Symbol} & \textbf{Description} \\ \midrule
  $\mathcal{D}_{orig}$ & Owner's original dataset without modifications \\
  $n$ & The number of samples  in $\mathcal{D}_{orig}$ \\
  $d$ & The number of features in $\mathcal{D}_{orig}$ \\
  $\mathcal{D}_{inj}$ & Synthesized samples to be injected into $\mathcal{D}_{orig}$ \\
  $\mathcal{D}_{mod}$ & Owner's modified dataset, i.e., $\mathcal{D}_{mod} = \mathcal{D}_{orig} \cup \mathcal{D}_{inj}$ \\ 
  $\mathcal{D}_{trig}$ & Synthesized dataset similar to $\mathcal{D}_{inj}$, which is used to query the target model for ownership detection \\ 
  $k$ & An integer value for numerical feature discretization \\
  $m$ & The number of pivots for data transformation \\
  $\rho$ & The injection ratio, i.e., $\rho = \num{\mathcal{D}_{inj}} / \num{\mathcal{D}_{orig}}$ \\
  $g$ & The number of selected empty balls \\ 
  $h$ & The number of synthetic samples per ball \\ 
  \bottomrule 
\end{tabular}
\vspace{0.5em}
\end{table}
\setlength{\textfloatsep}{1.0em}

\paragraph{Definition of Leaked Data Detection}
In this work, we investigate the problem of leaked data detection, which includes two parties: dataset owners and attackers.
The owner has a tabular dataset denoted by $\mathcal{D}_{orig}$, which is structured as a table comprising $n$ rows and $d$ feature columns, along with an additional column designated for class labels. Consequently, each row in this table represents a distinct sample instance.
Attackers are assumed to have circumvented security measures and gained unauthorized access to the owner's dataset. 
They then train a classification model using the designated label column and deploy it as a service. This model can be developed through any data processing techniques, training procedures, and various model types.
The objective is to develop a reliable method that allows dataset owners to ascertain, in a black-box setting, whether a given model has been trained using their dataset. 
We also explore possible adaptive attacks, assuming that attackers know the proposed method and possess certain information, such as the class distribution and feature ranges from which $\mathcal{D}_{orig}$ is drawn. 
However, we assume that they lack more detailed insights, such as the distribution of features by class or local class distribution at any subspace.
This problem is challenging as dataset owners lack access to the internal parameters of the target model, and their interaction is restricted to submitting queries and receiving corresponding predicted labels.

\paragraph{Comparisons to Model Backdooring through Data Poisoning}
This problem closely resembles the objective of model backdooring through data poisoning \cite{adi2018turning, zhang2018protecting}.
We aim to transform the owner's original dataset, $\mathcal{D}_{orig}$, into a modified version, $\mathcal{D}_{mod}$, ensuring that a model trained on either dataset will exhibit discernible differences in its predictions. 
Given that we operate under a black-box verification setting, our means of differentiation is to compare the predicted labels of designated trigger samples.
Nevertheless, there are three pivotal differences between our problem and the problem of model backdooring via data poisoning:
\begin{enumerate}[nolistsep,leftmargin=24pt,label*=(\arabic*)]
  \item Our primary goal is to empower dataset owners to verify and claim ownership of their dataset through a model trained on it. 
  Unlike strategies that might degrade model performance or embed malicious backdoors, the proposed method should have a minimal impact on both the dataset and the models trained on the modified version.

  \item Distinct from model backdooring, this research investigates a defense scenario where defenders do not require any knowledge or access to the attacker's target model. 
  Defenders can only interact with the model through model querying once the attacker has completed training. 
  Thus, the proposed solution should not make assumptions or modify the model's architecture, type, or training procedures.

  \item In the context of model backdooring through data poisoning, the poisoned dataset is typically inaccessible to parties other than the attackers, making it extremely difficult to detect or locate the backdoor in the trained model. 
  However, for the problem at hand, we assume that attackers have complete access to the modified dataset. This necessitates that any modifications to the dataset should be as discreet as possible to avoid detection and reversal by adversaries.
\end{enumerate}

These key differences render direct application of existing model backdooring and data poisoning solutions problematic, including random label flipping \cite{adi2018turning, zhang2018protecting} and training-time model manipulations like manipulating decision boundaries or parameter distributions \cite{le2020adversarial}. 
The former is susceptible to detection as outliers, while the latter assumes complete control of the training procedure.

\paragraph{Criteria for An Ideal Solution}
To address the first challenge, the proposed solution should abstain from altering samples in $\mathcal{D}_{orig}$. 
It must be crafted to minimize the impact on model accuracy for unseen samples drawn from the same population as $\mathcal{D}_{mod}$. 
To overcome the second and third challenges, the proposed method should be agnostic to the attackers' model training protocols. 
Furthermore, the injected samples should be meticulously designed to closely mirror the original dataset, thereby enhancing stealth and reducing the risk of detection.
Next, we introduce a novel approach, \textsc{LDSS}, which fulfills these stipulated criteria.
\section{LDSS: The Methodology}
\label{sect:method}

\subsection{Overview}
\label{sect:method:overview}

\textsc{LDSS} allows data owners to inject a series of synthetic samples, denoted as $\mathcal{D}_{inj}$, into their original dataset, $\mathcal{D}_{orig}$. 
This strategy is designed to deliberately shift the local class distribution. 
Once this injection is made, it is assumed that attackers would only have access to the modified dataset $\mathcal{D}_{mod}$.
For samples similar to the injected ones, any target model trained on $\mathcal{D}_{mod}$ is inclined to yield predictions distinct from model trained on $\mathcal{D}_{orig}$.
Thus, to discern whether datasets are leaked, data owners can generate a trigger set $\mathcal{D}_{trig}$ from similar distributions as $\mathcal{D}_{inj}$ for model querying.

\begin{figure}[t]%
\centering%
\captionsetup{skip=0.5em,belowskip=0em}%
\includegraphics[width=0.95\columnwidth]{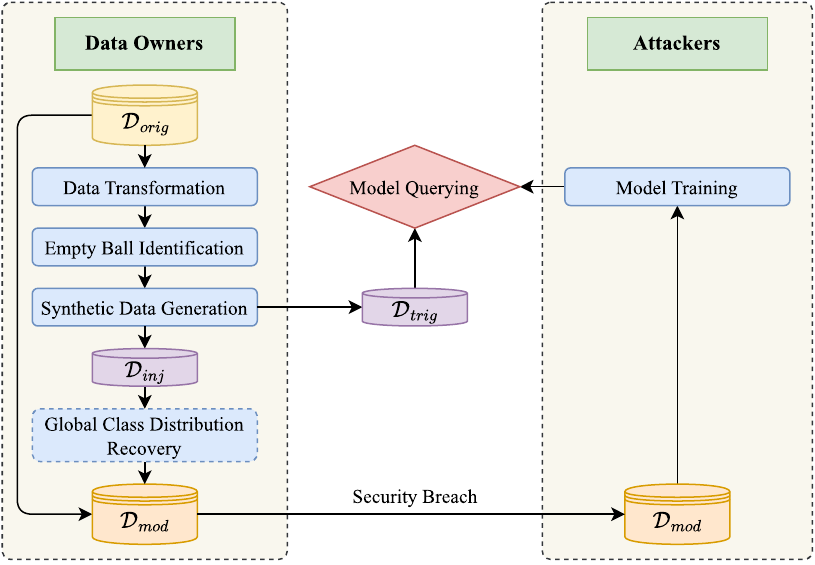}%
\caption{Overall flow of \textsc{LDSS}.}%
\label{fig:flow}%
\end{figure}%

Figure \ref{fig:flow} illustrates the overall flow of \textsc{LDSS}. 
To maintain the model's efficacy on genuine data resembling the distribution of $\mathcal{D}_{orig}$, \textsc{LDSS} needs to identify large empty balls that are devoid of any samples from $\mathcal{D}_{orig}$ before synthetic data generation.
Nevertheless, identifying large empty spaces in tabular datasets is challenging due to their mix of numerical and categorical features. 
To address this, we undertake a data transformation process, converting the dataset into an entirely numerical format.
When generating injection samples $\mathcal{D}_{inj}$, \textsc{LDSS} synthesize samples within a target class with the most significant frequency disparities relative to the prevailing local class. 
This strategy can shift the local class distribution.
However, it might also drastically alter the global class distribution and become discernible to attackers. 
To remedy this issue, we could perform an additional step to recover the global class distribution. 
Details are presented in the following subsections.

\subsection{Data Transformation}
\label{sect:method:transform}

\textbf{Challenges.}
To convert the tabular dataset to a wholly numerical format, a prevalent strategy is to apply one-hot encoding to categorical columns \cite{chawla2002smote}. 
Nevertheless, this strategy can result in extremely high dimensionality, especially when there are a lot of categorical features, each with a multitude of unique values. 
Not only does computing an empty ball become considerably slower, but such space also becomes very sparse. Thus, the empty ball no longer provides a good locality guarantee. 
Dimension reduction methods such as Principal Component Analysis (PCA) \cite{tipping1999mixtures, minka2000automatic, halko2011finding} are unsuitable here since the reduced space should be transformed back to the original space, which contains one-hot encoded features. 

To address the aforementioned challenges, we first convert the tabular dataset to an entirely categorical format. This allows us to employ a distance metric, e.g., Jaccard distance, to gauge the proximity between any pair of data. 
Subsequently, we determine $m$ representative pivots and utilize the Jaccard distance to these $m$ pivots as the numerical representation for each data.
The data transformation process encompasses three steps: numerical feature discretization, feature mapping, and numerical representation using pivots. 
Each of these steps is elucidated as follows.

\paragraph{Step 1: Numerical Feature Discretization}
\textsc{LDSS} first discretizes each numerical feature into $(k+1)$ bins using one-dimensional $k$-means clustering. 
Each bin is assigned a nominal value ranging from $0$ to $k$, in ascending order of their center values.
Here, the value of $k$ is chosen to accurately represent the distribution of the feature's values. 
To avoid inefficiencies in subsequent steps, $k$ is typically set to a moderate value, such as 5 or 10.

\paragraph{Step 2: Feature Mapping}
After the numerical feature discretization, we map each data $\bm{x} = (x_1, \cdots, x_d)$ into a set $\bm{\tilde{x}}$ that contains $(k \cdot d)$ elements, with each feature being represented by $k$ elements. This mapping is guided by Equations \ref{eqn:settransform} and \ref{eqn:settransform_2} as follows:
\begin{equation}\label{eqn:settransform}
  \bm{\tilde{x}} = \textstyle \bigcup_{i=1}^{d} T(i, x_i),
\end{equation}
where
\begin{equation}\label{eqn:settransform_2}
  T(i, x_i) =
  \begin{cases}
    \{(i, x_i, j) \mid j \in [1, k]\}, & \text{$i$ is categorical}; \\
    \{(i, x_i + j) \mid j \in [1, k]\}, & \text{$i$ is numerical}.
  \end{cases}
\end{equation}

For each categorical feature, identical values yield the same set of $k$ elements, ensuring that distinct values always produce $k$ distinct elements.
For any two numerical values, the number of identical elements among the $k$ elements reflects the closeness of their numerical values.
For example, if their categories after discretization are $v_1$ and $v_2$, then exactly $k-|v_1-v_2|$ elements will be the same after the feature mapping. 
The smaller the difference between the original values, the more elements they share after feature mapping. 
This mapping maintains the degree of similarity between two samples, and such similarity is scaled for numerical features to capture their distributions better.

\paragraph{Step 3: Numerical Representation using Pivots}
Finally, we determine $m$ sets as pivots and transform each set $\bm{\tilde{x}}$ into an $m$-dimensional vector $\bm{\hat{x}} = (\hat{x}_1, \cdots, \hat{x}_m) \in [0,1]^m$, where each coordinate $\hat{x}_i$ is represented by the Jaccard distance between $\bm{\tilde{x}}$ and $i$-th pivot. 
For any two sets $\bm{\tilde{x}}$ and $\bm{\tilde{y}}$, the Jaccard distance is defined as:
\begin{equation}\label{eqn:jaccarddist}
  J(\bm{\tilde{x}}, \bm{\tilde{y}}) = 1 - \frac{\num{\bm{\tilde{x}} \cap \bm{\tilde{y}}}}{\num{\bm{\tilde{x}} \cup \bm{\tilde{y}}}}.
\end{equation}
When selecting a small value for $m$, such as 5 or 10, the resulting space has a low dimension, facilitating the efficient identification of empty balls. 
Moreover, as the Jaccard distance is a distance metric that can effectively gauge the similarity between samples and pivots, samples in close proximity within the transformed space remain similar in the original feature space.

The next challenge is how to determine the pivots to preserve the proximity of the dataset.
An initial approach might be to randomly select pivots. 
In this method, pivots are drawn uniformly at random from $\mathcal{D}_{orig}$. 
However, this can lead to the selection of suboptimal pivots, especially if certain features in the randomly selected pivots are uncommon. As such, the Jaccard distance between the samples and such pivots can be consistently large.
A more refined strategy is to construct pivots based on value frequency.
Specifically, for each feature, we identify the top $m$ frequent values. These values are then assigned to the pivots in a \emph{round-robin} manner. If a feature has fewer than $m$ distinct values, we repeatedly use those values until every pivot has its set of values.
This strategy sidesteps the emphasis on infrequent values and also prevents the repetition of unique values, especially when a particular value is overwhelmingly common in specific features. Additionally, this strategy often results in more orthogonal pivots, especially when every feature has at least $m$ unique values.
In practice, $m$ can conveniently be set equal to $k$, which is the maximum number of discretized values of each numerical feature so that pivots include all the discretized values of numerical features.

\begin{table}[b]
\centering
\captionsetup{skip=0.5em}
\caption{Index mapping of transformed elements.}
\label{tab:index_mapping}
\begin{tabular}{p{0.12\linewidth}p{0.18\linewidth}} \toprule
  \textbf{Index} & \textbf{Element} \\ \midrule
  1  & (1, F, 1)   \\
  2  & (1, F, 2)   \\
  3  & (1, M, 1)   \\
  4  & (1, M, 2)   \\
  5  & (2, Cat, 1) \\
  6  & (2, Cat, 2) \\
  7  & (2, Dog, 1) \\
  8  & (2, Dog, 2) \\
  9  & (2, Fox, 1) \\
  10 & (2, Fox, 2) \\
  11 & (3, B, 1)   \\
  12 & (3, B, 2)   \\
  13 & (3, W, 1)   \\
  \bottomrule
\end{tabular}
\hspace{1em}
\begin{tabular}{p{0.12\linewidth}p{0.18\linewidth}} \toprule
  \textbf{Index} & \textbf{Element} \\ \midrule
  14 & (3, W, 2)   \\
  15 & (3, Y, 1)   \\
  16 & (3, Y, 2)   \\
  17 & (4, A, 1)   \\
  18 & (4, A, 2)   \\
  19 & (4, C, 1)   \\
  20 & (4, C, 2)   \\
  21 & (4, G, 1)   \\
  22 & (4, G, 2)   \\
  23 & (5, 1)      \\
  24 & (5, 2)      \\
  25 & (5, 3)      \\
  26 & (5, 4)      \\
  \bottomrule
\end{tabular}
\end{table}

\begin{figure*}[t]%
\centering%
\captionsetup{skip=0em,belowskip=0em}%
\subfigure[The pipeline of data transformation.]{%
  \label{fig:sample_transformation}%
  \includegraphics[width=0.88\textwidth]{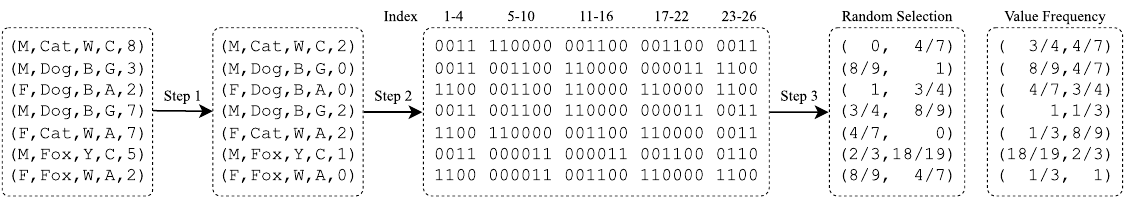}} \\%
\subfigure[Sample representations by random selection.]{%
  \label{fig:transform_random}%
  \includegraphics[width=0.44\textwidth]{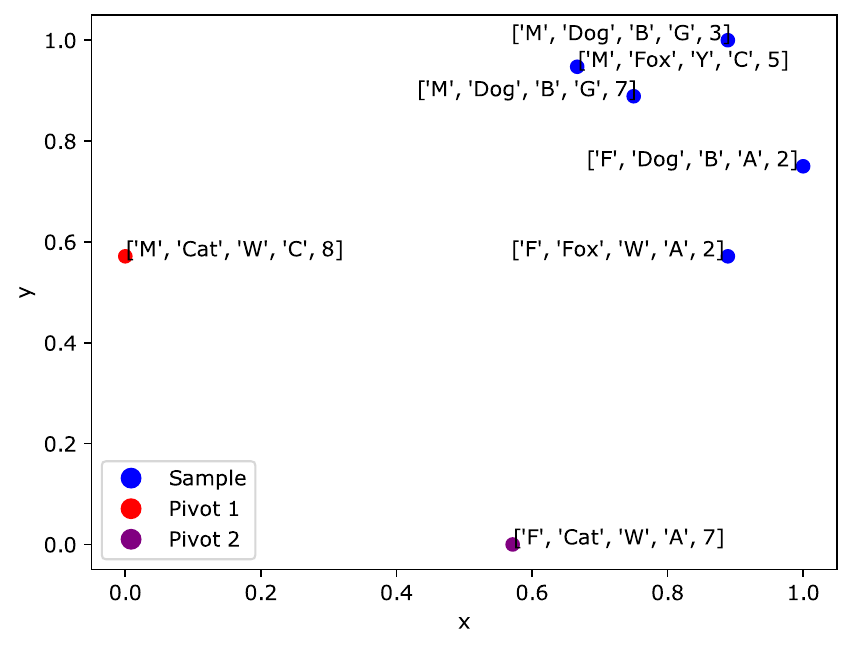}}%
\subfigure[Sample representations by value frequency.]{%
  \label{fig:transform_maxfreq}%
  \includegraphics[width=0.44\textwidth]{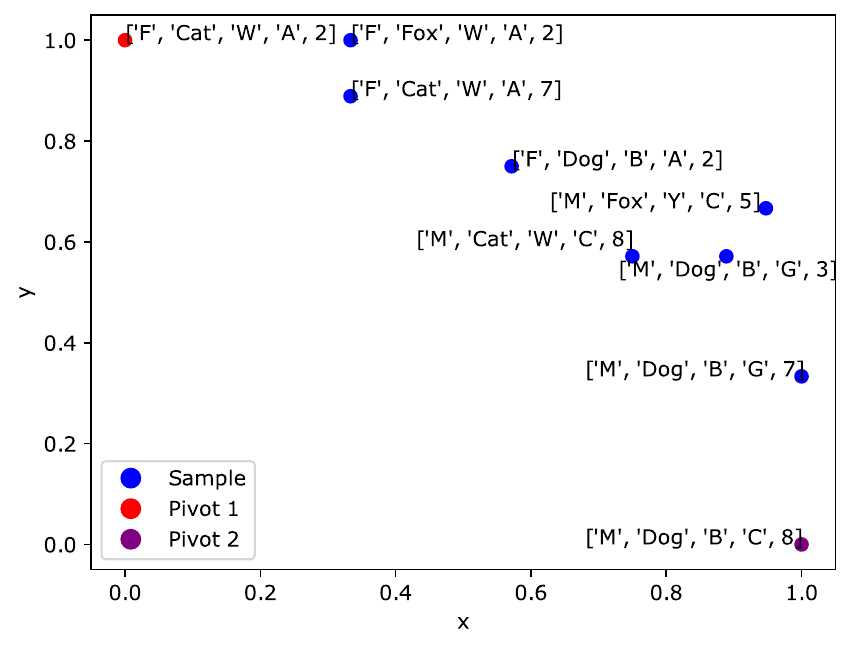}}%
\caption{Data transformation results using different pivot selection methods. Note that in Figure \ref{fig:transform_random}, both pivots come from original samples, while in Figure \ref{fig:transform_maxfreq}, both pivots are constructed based on value frequency. In Figure \ref{fig:sample_transformation}, we present the final results using random selection and value frequency, which map to Figures \ref{fig:transform_random} and \ref{fig:transform_maxfreq}, respectively.}%
\label{fig:transformation}%
\end{figure*}%

\begin{example}\label{example:transformation}
  Suppose a small pet dataset contains seven samples, as illustrated at the beginning of Figure \ref{fig:sample_transformation}. 
  This dataset contains $d=5$ feature columns, namely gender, species, code of color, code of the country of origin, and age. The first four features are categorical, while the last one is numerical. Let $m=k=2$. 
  
  In step 1, the last numerical feature is discretized into $k + 1 = 3$ bins, with values 2 and 3 in bin 0, 5 in bin 1, and 7 and 8 in bin 2. For example, the first sample becomes $(M, Cat, W, C, 2)$.
  In step 2, each sample is mapped to $k d = 10$ elements. Taking the first sample as an example, it maps to the set: $\{(1, M, 1)$, $(1, M, 2)$, $(2, Cat, 1)$, $(2, Cat, 2)$, $(3, W, 1)$, $(3, W, 2)$, $(4, C, 1)$, $(4, C, 2)$, $(5, 3)$, $(5, 4)\}$. 
  To streamline our explanation, elements after feature mapping are uniquely indexed, as shown in Table \ref{tab:index_mapping}. 
  Furthermore, in Figure \ref{fig:sample_transformation}, we employ one-hot encoding to transform sets of elements into binary strings. Thus, the first sample is represented as 0011~110000~001100~001100~0011.
  In step 3, we select two pivots ($k=2$) and compute the Jaccard distance between samples and pivots. 
  For the approach using random selection, e.g., selecting the first and the fifth samples as pivots, the resulting representation is shown in the left column of Figure \ref{fig:sample_transformation} following Step 3.
  For the approach based on value frequency, e.g., constructing a pair of pivots $(F, Cat, W, A, 2)$ and $(M, Dog, B, C, 8)$--with the last feature discretized to $0$ and $2$, respectively--the resulting representation is illustrated in the right column of Figure \ref{fig:sample_transformation} after Step 3. 

  In Figure \ref{fig:transform_random} and \ref{fig:transform_maxfreq}, we depict the sample representations of both approaches using scatter plots.
  These figures highlight that pivots built on value frequency more effectively preserve the relative similarities among samples post-transformation compared to those by random selection. 
  This is because the random pivot selection might incorporate rare or duplicate values for specific features, causing many unique values to become indistinguishable.
  On the other hand, pivots derived based on value frequency tend to be more orthogonal, thus better maintaining relative similarity.
  \hfill $\triangle$ \par 
\end{example}

\subsection{Empty Ball Identification} 
\label{sect:method:empty_ball}

The next step is identifying empty balls in the transformed hypercube $[0,1]^m$. 
Our objective is to locate those empty balls that are both sufficiently large and close to the transformed dataset $\widehat{\mathcal{D}}_{orig}$. Balls located too far from this dataset are considered outliers, making them unsuitable for synthetic data injection.
Before delving into our strategy to identify these empty balls, we first define an empty ball for numerical features.

\paragraph{Definition of the Empty Ball}
Let $\bm{\hat{c}} = (\hat{c}_1, \cdots, \hat{c}_m)$ represent the center of an empty ball in the $m$-dimensional transformed space. 
To define the radius of this empty ball, we consider the prevalent Euclidean distance between any two points $\bm{\hat{x}} = (\hat{x}_1, \cdots, \hat{x}_m)$ and $\bm{\hat{y}} = (\hat{y}_1, \cdots, \hat{y}_m)$, i.e., $\norm{\bm{\hat{x}} - \bm{\hat{y}}} = \textstyle \sqrt{\sum _{i=1}^{m} (\hat{x}_i - \hat{y}_i)^2}$.
With this distance, the radius $r$ of the empty ball centered at $\bm{\hat{c}}$ is defined below:
\begin{equation}\label{eqn:emtpy_ball}
  r = \min_{\bm{\hat{x}} \in \widehat{\mathcal{D}}_{orig}} \norm{\bm{\hat{x}}-\bm{\hat{c}}}.
\end{equation}

\paragraph{Empty Ball Identification}
To prevent the over-clustering of samples generated within the empty balls, which could allow attackers to easily identify injected samples, we prioritize the balls with larger radii over those with smaller ones.
Several heuristics, such as the Voronoi diagram and the Evolutionary Algorithm, are available for finding the largest empty ball, as discussed by \citet{lee2004stochastic}.
However, our goal is merely to identify large empty balls. 
For enhanced efficiency, we adopt Simulated Annealing (SA) coupled with constrained updates. The procedure is detailed in Algorithm \ref{alg:sphere-gen}.

The algorithm initiates by randomly selecting a batch of $G$ samples (500, in our experiments) from $\widehat{\mathcal{D}}_{orig}$ as seeds (potential centers for empty balls) (Line \ref{ball:init}). 
Then, it continues to replace these seeds with perturbed versions that have a greater radius (Lines \ref{ball:iter-start}--\ref{ball:iter-end}). 
As iterations progress, the perturbation scale is reduced, governed by the temperature parameter $t$ (Line \ref{ball:reduce-t}). 
Given an arbitrary center $\bm{\hat{c}} = (\hat{c}_1, \cdots, \hat{c}_d)$,
the function $perturb(\bm{\hat{c}},t)$ adds noise that is scaled by $t$ and the standard deviation $std_i$ of each dimension $i$ over all samples, i.e., $\hat{c}_i^\prime = \hat{c}_i + t \cdot uniform(-std_i, std_i)$.
The function $r(\bm{\hat{c}})$ is defined by Equation \ref{eqn:emtpy_ball}, computing the radius of the empty ball centered at $\bm{\hat{c}}$.
In each iteration, ball centers are updated by the perturbed ones that have a larger radius (Line \ref{ball:stop}).
To ensure that the empty balls remain in proximity to $\widehat{\mathcal{D}}_{orig}$ and avoid positioning synthesized samples as potential outliers relative to $\widehat{\mathcal{D}}_{orig}$, only ball centers that are not flagged as outliers by Isolation Forest \cite{liu2008isolation}, trained on $\widehat{\mathcal{D}}_{orig}$, are considered (Line \ref{ball:stop}). 
These selected centers then supersede the prior ones (Line \ref{ball:update-center}).
The entire process terminates when $t$ is less than a predefined threshold $\epsilon$ (Line \ref{ball:termination}).

\begin{algorithm}[t]
\small
\caption{Empty Ball Identification}
\label{alg:sphere-gen}
\KwIn{Transformed dataset $\widehat{\mathcal{D}}_{orig} \subset [0,1]^m$, the number of empty balls $G$, stopping condition $\epsilon$;}
\KwOut{A set of empty balls $\mathcal{S}$;}
$S \gets G$ samples drawn uniformly at random from $\widehat{\mathcal{D}}_{orig}$\; \label{ball:init}
$t \gets 1$\;
\While{$t \geq \epsilon$}{ \label{ball:termination}
  \ForEach{$\bm{\hat{c}} \in S$}{ \label{ball:iter-start}
    \Do{$\bm{\hat{c}}^\prime$ is an outlier or $r(\bm{\hat{c}}^\prime) \leq r(\bm{\hat{c}})$}{
      $\bm{\hat{c}}^\prime \gets perturb(\bm{\hat{c}},t)$\;
    } \label{ball:stop}
    $\bm{\hat{c}} \gets \bm{\hat{c}}^\prime$\; \label{ball:update-center}
  } \label{ball:iter-end}
  $t \gets t \cdot 0.8$\; \label{ball:reduce-t}
}
\Return $S$\;
\end{algorithm}

As an adequately large empty ball suffices for generating distant samples, a moderate decrement of $t$ in every iteration, coupled with a small value of $\epsilon$, can ensure good performance without compromising execution speed.
In the experiments, we set $\epsilon = 0.01$ and decrease $t$ by a factor of 0.8 in each iteration.

\subsection{Synthetic Data Injection}
\label{sect:method:synthetic}

Upon identifying $G$ large empty balls, we proceed with two hyperparameters, $g$ ($1 \leq g \leq G$) and $h$ for synthetic data injection. This process comprises three steps: empty ball selection, synthetic data generation, and synthetic data injection.

\paragraph{Step 1: Empty Ball Selection}
To produce high-quality synthetic data, our goal is to identify empty balls with a substantial local frequency gap. 
We first pinpoint the top $g$ empty balls that exhibit the most pronounced local frequency disparities between the highest and lowest frequency classes.
Then, using the least frequent class as our target class, we generate $h$ synthetic samples of this class within each of these chosen balls.
The local frequency gap is ascertained by counting frequencies within the closest $h$ samples to the ball's center based on Euclidean distance. 
This implies that within the local vicinity of $2h$ samples centered around the ball--post the synthetic data injection--the target class will emerge as the dominant one since at least $h$ of the injected samples will bear the same class.
Due to the use of SA, Algorithm \ref{alg:sphere-gen} might produce near-duplicated empty balls. Thus, duplicate removal is performed by shrinking the radius of balls with lower ranks.

Given the ranking criteria for empty balls, we ensure that a minority class takes precedence locally. 
Thus, this is anticipated to affect the decision-making of a classification model within those localized regions.
We typically set $g$ at a small value (e.g., 10) to concentrate synthetic samples. $h$ is then determined by $g$ and the injection ratio $\rho$, where $\rho = \num{\mathcal{D}_{inj}} / \num{\mathcal{D}_{orig}}$. 


\paragraph{Step 2: Synthetic Data Generation}
After identifying the top $g$ empty balls, where each is assigned a designated target class, \textsc{LDSS} synthesizes $h$ samples per ball to form an injection set. 
Since the empty balls exist in the transformed space, the resulting samples should be transformed into the original feature space. 
It is important to note that the direct random generation of samples per empty ball, followed by their transformation back to the original feature space, is not feasible. 
The primary reason is the inherent challenge (or the sheer impossibility) of finding a sample that perfectly matches the target Jaccard distances to all $m$ pivots.

For a dataset with $d$ features, we map each feature into $k$ elements based on Equations \ref{eqn:settransform} and \ref{eqn:settransform_2}.
If they share $l$ common elements, their Jaccard distance is given by $J = 1 - \tfrac{l}{2dk-l}$. 
Given a Jaccard distance $J$, the number of shared elements can be deduced as Equation \ref{eqn:l}:
\begin{equation}\label{eqn:l}
  l = l(J) = \tfrac{2dk(1-J)}{2-J}.
\end{equation}
Considering a distinct value $v'$ of feature $i$ in $\mathcal{D}_{orig}$, let $\bm{w}_{v'}$ be an $m$-dimensional vector, where its coordinates are the number of shared elements between $v'$ and the $i$-th feature of $m$ pivots.
Hence, we can calculate the total number of shared elements between the $m$ pivots and a certain sample projected to $\bm{\hat{c}} = (\hat{c}_1, \cdots, \hat{c}_m)$ as $\bm{w}_{\bm{\hat{c}}} = (w_1, \cdots, w_m)$ such that $w_j = l(\hat{c}_j)$, where $j \in \{1, \cdots, m\}$. 
Thus, by utilizing the vector $\bm{w}_{v'}$ and the target vector $\bm{w}_{\bm{\hat{c}}}$ as a measure of weights, we can reduce this problem to a classical $m$ dimensional knapsack problem \cite{knuth2014art}. 

Consider a bag of infinite capacity and $m$ dimensional weights and a target weight $\bm{w}_{\hat{\bm{c}}}$ to achieve by filling the bag with selected objects. 
Each available object is identified by a pair ($i,v^\prime$) representing feature $i$ of value $v^\prime$, and its weight is thus $\bm{w}_{v^\prime}$. 
In this knapsack problem, the weight vector $\bm{w}_{v^\prime}$ is non-negative in all $m$ dimensions. 
There exists one such object for each $i \in \{1, \cdots, d\}$ and possible values $v$ of feature $i$. 
A total of $d$ objects should be chosen with exactly one for each $i$. 
The objective is to fill the bag with the chosen $d$ objects such that the total weight $\bm{w}$ is close to $\bm{w}_{\bm{\hat{c}}}$ measured by $L_\infty$ distance. 
Each possible solution consists of exactly $d$ objects representing the values for each feature. 
Thus, we employ dynamic programming to solve the knapsack problem and synthesize samples as a solution to our original problem. The pseudo-code is presented in Algorithm \ref{alg:DPSync}.

Let state $S_{i,\bm{w},v}$ be a Boolean value denoting whether the bag can be filled by $i$ items from the first $i$ features, with $v$ as feature $i$'s value and a total weight $\bm{w}$. 
Algorithm \ref{alg:DPSync} finds all possible $S_{i,\bm{w},v}$ for each $1 \leq i \leq d$ (Lines \ref{dp:C-start}--\ref{dp:C-end}). 
We then choose among all $S_{d,\bm{w},v}=True$ whose $\bm{w}$ has the smallest $L_\infty$ distance to $\bm{w}_{\bm{\hat{c}}}$ (Line \ref{dp:target}). 
Finally, we backtrack to generate synthetic samples in the original $d$ dimensional space with random values selected for a $v$ if there are different possibilities (Lines \ref{dp:backtrack_start}--\ref{dp:backtrack_end}) and return $R$ as the result set (Line \ref{dp:end}).

It is worth noting that directly executing Algorithm \ref{alg:DPSync} is practically challenging due to the potentially vast number of possible triples $(i,\bm{w},v)$ that $S_{i,\bm{w},v} = True$, especially considering the exponential growth of samples constructed from every possible feature value. 
We employ an effective approximation by capping the number of possible states for each $i$ at the end of the loop (Lines \ref{dp:C-start}--\ref{dp:C-end}). 
For each $i$, we only keep a limited number of $S_{i,\bm{w},v}=True$ that $\bm{w}$ has smallest $L_\infty$ distance to $\bm{w}_{\bm{\hat{c}}} \cdot \tfrac{i}{d}$ by enforcing a similar weight increase rate in all dimensions. 
In practice, we observe that such a limit between 50,000 and 100,000 is satisfactory. 
Other optimization includes memorizing only true states and enumerating only at most $(m+1)$ possible feature values instead of all (Line \ref{dp:possible-feature}).

\begin{algorithm}[t]
\small
\caption{Sample Synthesis within An Empty Ball}
\label{alg:DPSync}
\KwIn{$\mathcal{D}_{orig}$ and $m$ pivots with $d$ features, an empty ball center $\bm{\hat{c}} \in [0,1]^m$ and its target weight vector $\bm{w}_{\bm{\hat{c}}}$, and the number of synthetic samples per ball $h$;}
\KwOut{Result set $R$;}
$S_{0,[0]^m,nil} = True$\; \label{dp:C-init}
\For{$i = 1$ \KwTo $d$} { \label{dp:C-start}
  \ForEach{$(\bm{w},v)$ that $S_{i-1,\bm{w},v} = True$}{
    \ForEach{$v'$ in distinct values of feature $i$ in $\mathcal{D}_{orig}$} { \label{dp:possible-feature}
      $\bm{w}_{v^\prime} \gets$ \# shared elements between $v^\prime$ and the $i$-th feature of $m$ pivots\;
      $\bm{w}^\prime \gets \bm{w} + \bm{w}_{v^\prime}$\;
      $S_{i,\bm{w}^\prime,v^\prime} = True$\;
    }
  }
} \label{dp:C-end}
$T \gets$ $h$ pairs $\{(\bm{w}, v)\}$ s.t. $S_{d,\bm{w},v} = True$ with smallest $\norm{\bm{w} -\bm{w}_{\bm{\hat{c}}}}_{\infty}$\; \label{dp:target}
$R \gets \varnothing$\; \label{dp:backtrack_start}
\ForEach{$(\bm{w}, v) \in T$} {
  $\bm{x}=(x_1,\cdots,x_d) \gets [0]^d$\;
  \For{$i = d$ \KwTo $1$} {
    $x_i \gets v$\;
    $\bm{w}_v \gets$ \# shared elements between $v$ and the $i$-th feature of $m$ pivots\;
    $\bm{w}^\prime \gets \bm{w} - \bm{w}_v$\;
    $v^\prime \gets $ a random value $v'$ that $S_{i-1,\bm{w}^\prime, v^\prime} = True$\;
    $(\bm{w}, v) \gets (\bm{w}^\prime, v^\prime)$\;
  }
  $R \gets R \cup \{\bm{x}\}$\;
}\label{dp:backtrack_end}
\Return $R$\; \label{dp:end}
\end{algorithm}

\paragraph{Step 3: Synthetic Data Injection}
After executing Algorithm \ref{alg:DPSync}, we find the top $h$ samples with the closest $L_\infty$ distance to the ball center. 
This is repeated for all $g$ empty balls to synthesize the total $(g \cdot h)$ samples as injection set $\mathcal{D}_{inj}$. After injecting them into  $\mathcal{D}_{orig}$, we get a modified dataset, i.e., $\mathcal{D}_{mod} = \mathcal{D}_{orig} \cup \mathcal{D}_{inj}$. 

Assuming that $\widehat{\mathcal{D}}_{orig}$ accurately represents the population, empty balls will either remain unoccupied or be minimally populated in datasets of similar distributions. 
Thus, this approach guarantees a considerable local distribution shift, even when confronted with a different dataset originating from the same population.
It also provides a defense against potential dilution attacks. 
Should an attacker acquire additional data from the same population and merge it with $\widehat{\mathcal{D}}_{orig}$ for training, the samples injected within these empty balls, which are likely to remain unoccupied in datasets from the same source, will continue to effectively alter the local distribution.

\begin{figure}[t]%
\centering%
\captionsetup{skip=0.5em,belowskip=0em}%
\includegraphics[width=0.3\columnwidth]{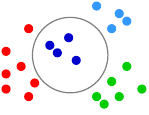}%
\caption{An example illustrates the injection of synthetic samples into an empty ball. By injecting four dark blue points with labels identical to the light blue ones, a model trained on this modified dataset $\mathcal{D}_{mod}$ is more likely to predict samples within the grey circle as blue rather than red.}%
\label{fig:empty-ball}%
\end{figure}

\begin{example}\label{example:empty-ball}
  Figure \ref{fig:empty-ball} visually conveys the idea of synthetic data injection. 
  The red, green, and light blue points depict the transformed data from 3 different classes in $\widehat{\mathcal{D}}_{orig}$. 
  The grey circle shows an empty ball among them, where the most frequent class is red, and the least frequent class is blue. 
  The model trained on $\widehat{\mathcal{D}}_{orig}$ would likely classify samples within the circle as red.
  Yet, when injecting 4 dark blue samples labeled similarly to the light blue ones into $\widehat{\mathcal{D}}_{orig}$, the model trained on $\mathcal{D}_{mod}$ would probably predict samples inside the grey circle as blue. 
  This distinction is crucial for determining whether a model was trained using $\mathcal{D}_{mod}$.
  \hfill $\triangle$ \par 
\end{example}

\subsection{Model Querying}
\label{sect:method:query}

To determine whether a target model is trained on $\mathcal{D}_{mod}$, we can form a trigger set $\mathcal{D}_{trig}$ by randomly selecting samples from $\mathcal{D}_{inj}$ and querying the model. 
However, this method is not optimal as attackers might deliberately alter predictions if they detect that the query samples are the same as those from $\mathcal{D}_{mod}$. 
A more robust approach involves using Algorithm \ref{alg:DPSync} to create $\mathcal{D}_{trig}$ with a different random seed. 
It is because feature $i$ can take on multiple values for a given state $S_{i,\bm{w},v}$. Thus, varying the random seed can generate different samples sharing the same state. 

For samples in $\mathcal{D}_{trig}$, models trained on $\mathcal{D}_{mod}$ are expected to yield predictions consistent with the original labels.
We assess the accuracy by comparing these predicted labels against the original ones.
If the accuracy exceeds a specific threshold $\tau$, we consider that the model was trained on $\mathcal{D}_{mod}$. 
Typically, $\tau$ can range widely between $[0.4, 0.8]$. This range is chosen because a model trained on $\mathcal{D}_{orig}$ or a dataset from a similar distribution usually has an accuracy close to 0, while the accuracy nears 1 for models trained on $\mathcal{D}_{mod}$. 
In practice, a large $\tau$ is usually preferred to prevent false alarms, even if it sacrifices the true positive rate.

\subsection{Global Class Distribution Recovery}
\label{sect:method:recovery}

There is a possibility that the majority of samples in $\mathcal{D}_{inj}$ share the same label, leading to a significant shift in the global class distribution.
The attackers, who have a rough idea about the class distribution in the general population, might exploit this to detect datasets modified by our algorithm or others of a similar nature. 

To bridge the class distribution gap, we slightly perturb some samples in $\mathcal{D}_{orig}$ and incorporate them into $\mathcal{D}_{inj}$. 
First, we determine the number of samples needed to restore each class's distribution. Then, samples are chosen uniformly at random. 
For every selected sample, numerical features are adjusted with a tiny noise (e.g., 5\%) based on their maximum value range, and there is a certain probability (e.g., 20\%) of changing a categorical feature's value, drawing from other samples with the same class. 
While this adjustment remains optional, we applied it across all \textsc{LDSS} modified datasets in the experiments before assessing the performance.

\subsection{Extension to Regression Tasks}
\label{sect:method:regression}

While \textsc{LDSS} has primarily been applied to classification tasks for detecting leaked datasets, its versatility allows for seamless adaptation to regression tasks. 
To facilitate this, we initially discretize the target column into a few categories to serve as class labels. \textsc{LDSS} is then employed accordingly, and the label of each injected and trigger sample is substituted with the target value from a randomly selected original dataset sample of the same class. 
We anticipate that \textsc{LDSS} will induce a significant difference in regression errors, such as Mean Absolute Error (MAE) or Mean Square Error (MSE), between models trained on $\mathcal{D}_{orig}$ and those on $\mathcal{D}_{mod}$. 
This will be demonstrated in Section \ref{sect:expt:regression}, underscoring \textsc{LDSS}'s versatility and effectiveness across different machine learning tasks.
\section{Experiments}
\label{sect:expt}

We systematically evaluate the performance of \textsc{LDSS} for leaked data detection to answer the following questions: 
\begin{itemize}[nolistsep,leftmargin=20pt]
  \item \textbf{Reliability}: Can \textsc{LDSS} detect leaked datasets based on the trigger prediction accuracy of the target model? (Section \ref{sect:expt:rely_and_robust})
  
  \item \textbf{Robustness}: How consistent is the detection accuracy across various machine learning models? (Section \ref{sect:expt:rely_and_robust}) 
  
  \item \textbf{Fidelity}: Does \textsc{LDSS} preserve prediction accuracy on testing samples at a comparable rate? (Section \ref{sect:expt:fidelity})
  
  \item \textbf{Security}: Are the injected samples stealthy enough to make detection and removal challenging? (Section \ref{sect:expt:security})  
  
  \item \textbf{Efficiency}: Does \textsc{LDSS} achieve its goals with a minimal injection size to reduce computational time? (Section \ref{sect:expt:efficiency})
  
  \item \textbf{Regression}: Is \textsc{LDSS} versatile enough to effectively support the regression task? (Section \ref{sect:expt:regression})
\end{itemize}

In addition, we include the parameter study in Section \ref{sect:expt:params}. Before delving into the results, we first present the experimental setup.

\subsection{Experimental Setup}
\label{sect:expt:setup}

\noindent\textbf{Datasets.}
We employ five publicly available, real-world datasets for performance evaluation, including
\begin{itemize}[nolistsep,leftmargin=20pt]
  \item \textbf{Adult},\footnote{\url{http://archive.ics.uci.edu/ml/datasets/adult}} derived from the 1994 Census database, has 2 classes based on whether a person earns above or below \$50K a year. 
  
  \item \textbf{Vermont} and \textbf{Arizona}\footnote{\url{https://datacatalog.urban.org/dataset/2018-differential-privacy-synthetic-data-challenge-datasets/resource/2478d8a8-1047-451b-ae23}} are US Census Bureau PUMS files for Vermont and Arizona states, respectively. We defined 4 classes from the wage column for classification: 0 (where wage $=0$), 1 (wage $\leq 500$), 2 (wage $\leq 1,000$), and 3 (wage $> 1,000$).

  \item \textbf{Covertype}\footnote{\url{http://archive.ics.uci.edu/ml/datasets/covertype}} compromises cartographic variables of a small forest area and associated observed cover types (with 7 forest cover types in total) \cite{Dua:2019}. 
  
  \item \textbf{GeoNames}\footnote{\url{https://www.kaggle.com/geonames/geonames-database}} covers over 11 million placenames. We retained only 8 attributes for classification, i.e., latitude, longitude, feature class, country code, population, elevation, dem, and timezone. The \emph{feature class} attribute serves as the target label.
\end{itemize}

To ensure a fair comparison and consistency, we exclude samples lacking class labels and features with over one-third missing values or full correlation with others. 
We address the remaining missing values through random imputation from a valid sample.

\paragraph{Compared Methods}
We compare the following methods for leaked data detection in the experiments. 
\begin{itemize}[nolistsep,leftmargin=20pt]
  \item \textbf{\textsc{LDSS}} is our method proposed in Section \ref{sect:method}. It identifies $g$ large empty balls and injects $h$ samples into each empty ball. 
  
  \item \textbf{\textsc{Flip}} is a baseline that chooses a set of $(g \cdot h)$ samples from $\mathcal{D}_{orig}$ and flips their labels uniformly at random. It directly adapts label flipping techniques such as \cite{adi2018turning, namba2019robust}.

  \item \textbf{\textsc{FlipNN}} enhances \textsc{Flip} by randomly selecting $g$ samples and flipping their $h$ nearest neighbors to the least frequent label.
  It is tailored to aid simpler models in learning locally modified information, as a single flipped sample in a local area might not be effective unless the model is complex and highly parameterized, like DNNs, for better memorization.
\end{itemize}

\textsc{FlipNN} can be considered a random version of \textsc{LDSS} without searching for large empty balls. For \textsc{Flip} and \textsc{FlipNN}, the flipped samples are directly used as trigger samples. 
Note that watermarking techniques are not our direct competitors, as our focus is on detecting leaked data, which requires no knowledge of the training process and is applicable across various models, not limited to DNNs.
Moreover, the goals of these watermarking methods significantly differ from ours.
Their approach of embedding data within DNNs is less suitable for tabular datasets, which do not exhibit the local feature correlations typical in images.
Additionally, tabular datasets often rely on simpler models like Decision Trees, which are less equipped for handling intricate watermark embedding.

\paragraph{Evaluation Measures}
We adopt evaluation criteria from model watermarking \cite{darvish2019deepsigns}, focusing on detecting leaked data. 
Given our black-box scenario without insights into training or model architecture, our assessment concentrates on prediction outcomes when querying the target model.
We gauge the detection accuracy and standard deviation of trigger samples across seven conventional classification models: Naive Bayes (NB), $k$-Nearest Neighbor ($k$-NN), Linear Support Vector Classifier (LSVC), Logistic Regression (LR), Multi-Layer Perceptron (MLP), Decision Tree (DT), and Random Forest (RF).
Ideally, models trained on $\mathcal{D}_{orig}$ should have low trigger accuracy, while those on $\mathcal{D}_{mod}$ should score higher.

\paragraph{Experiment Environment}
All methods were written in Python 3.10. All experiments were conducted on a server with 24 CPUs of Intel\textsuperscript{\textregistered} Xeon\textsuperscript{\textregistered} E5-2620 v3 CPU @ 2.40 GHz, 64 GB memory, and one NVIDIA GeForce RTX 3090, running on Ubuntu 20.04.

\begin{table*}[t]
\centering
\captionsetup{skip=0.5em}
\setlength\tabcolsep{5pt}
\small
\caption{Training and testing average Accuracy (Acc, \%), Average Absolute Difference (AAD, \%), and Maximum Absolute Difference (MAD, \%) of classification models trained on the original dataset and the modified dataset by \textsc{Flip}, \textsc{FlipNN} and \textsc{LDSS} 
(Note: The training (testing) AAD of each method is larger or equal to the absolute difference between average training (testing) Acc. of Original and its method because AAD is calculated as the mean of their absolute differences among 11-fold runs).}
\label{tab:fidelity}
\resizebox{\textwidth}{!}{%
  \begin{tabular}{ccccccccccccccccccccc} \toprule
  \multirow{2}[6]{*}{\textbf{Datasets}} & \multicolumn{2}{c}
  {\textbf{Original}} & \multicolumn{6}{c}
  {\textbf{\textsc{Flip}}} & \multicolumn{6}{c}{\textbf{\textsc{FlipNN}}} & \multicolumn{6}{c}{\textbf{\textsc{LDSS}}} \\
  \cmidrule(lr){2-3}\cmidrule(lr){4-9}\cmidrule(lr){10-15}\cmidrule(lr){16-21}  &{\textbf{Training}} &{\textbf{Testing}} & \multicolumn{3}{c}{\textbf{Training}} & \multicolumn{3}{c}{\textbf{Testing}} & \multicolumn{3}{c}{\textbf{Training}} & \multicolumn{3}{c}{\textbf{Testing}} & \multicolumn{3}{c}{\textbf{Training}} & \multicolumn{3}{c}{\textbf{Testing}} \\ 
  \cmidrule(lr){2-2} \cmidrule(lr){3-3} \cmidrule(lr){4-6} \cmidrule(lr){7-9} \cmidrule(lr){10-12} \cmidrule(lr){13-15} \cmidrule(lr){16-18} \cmidrule(lr){19-21} & \textbf{Acc} & \textbf{Acc} & \textbf{Acc} & \textbf{AAD} & \textbf{MAD} & \textbf{Acc} & \textbf{AAD} & \textbf{MAD} & \textbf{Acc} & \textbf{AAD} & \textbf{MAD} & \textbf{Acc} & \textbf{AAD} & \textbf{MAD} & \textbf{Acc} & \textbf{AAD} & \textbf{MAD} & \textbf{Acc} & \textbf{AAD} & \textbf{MAD} \\ 
  \midrule
    Adult     & 84.63 & 83.60 & 77.83 & 6.80 & 7.72 & 82.96 & 0.64 & 1.33 & 78.05 & 6.58 & 7.77 & 80.91 & 2.69 & 6.39 & 86.61 & 1.98 & 3.23 & 83.07 & 0.77 & 2.00 \\
    Vermont   & 75.29 & 72.14 & 67.90 & 7.39 & 7.75 & 71.67 & 0.59 & 1.24 & 68.60 & 6.69 & 7.76 & 68.53 & 3.61 & 9.37 & 77.71 & 2.42 & 4.44 & 71.06 & 1.08 & 2.30 \\
    Arizona   & 80.34 & 77.93 & 72.34 & 8.00 & 8.33 & 77.42 & 0.53 & 1.61 & 73.95 & 6.39 & 8.85 & 73.94 & 3.99 & 9.18 & 82.02 & 2.75 & 3.59 & 76.72 & 1.37 & 3.68 \\
    Covertype & 73.05 & 72.52 & 67.58 & 5.47 & 7.02 & 72.30 & 0.37 & 0.77 & 70.38 & 2.85 & 4.14 & 68.63 & 3.89 & 6.47 & 72.54 & 1.79 & 3.73 & 70.68 & 1.87 & 4.57 \\
    GeoNames  & 56.43 & 56.35 & 51.65 & 4.78 & 5.31 & 56.32 & 0.05 & 0.13 & 52.97 & 3.46 & 4.12 & 54.37 & 1.98 & 4.83 & 59.23 & 2.80 & 3.72 & 55.54 & 0.81 & 2.56 \\
  \bottomrule
  \end{tabular}
}
\end{table*}

\paragraph{Parameter Setting}
In the experiments, we set $\rho = 10\%$ and $g=10$ and configure the contamination for Isolation Forest training at 0.05 for empty ball identification. 
We assess the outlier percentage of injected samples at contamination levels of 0.01, 0.05, and 0.1.
Both $m$ and $k$ are fixed at 10. Yet, for features with limited distinct values, the actual number of discretized values is less than $k$. 

For each experiment, 11-fold cross-validation is conducted with a fixed random seed, where each round utilizes one fold as the training set and the next as the testing set. 
We employ \textsc{LDSS} to generate both injection and trigger sets and evaluate the classification models trained on the training set with or without the injection set. 
Afterward, we incrementally incorporate the remaining nine folds of data into the training set. 
We repeat these evaluations on those amalgamated datasets. This allows us to investigate the effectiveness of \textsc{LDSS} under the potential threat of dilution attacks.

\begin{figure}[t]%
\centering%
\captionsetup{skip=0em,belowskip=0em}%
\includegraphics[width=0.95\columnwidth]{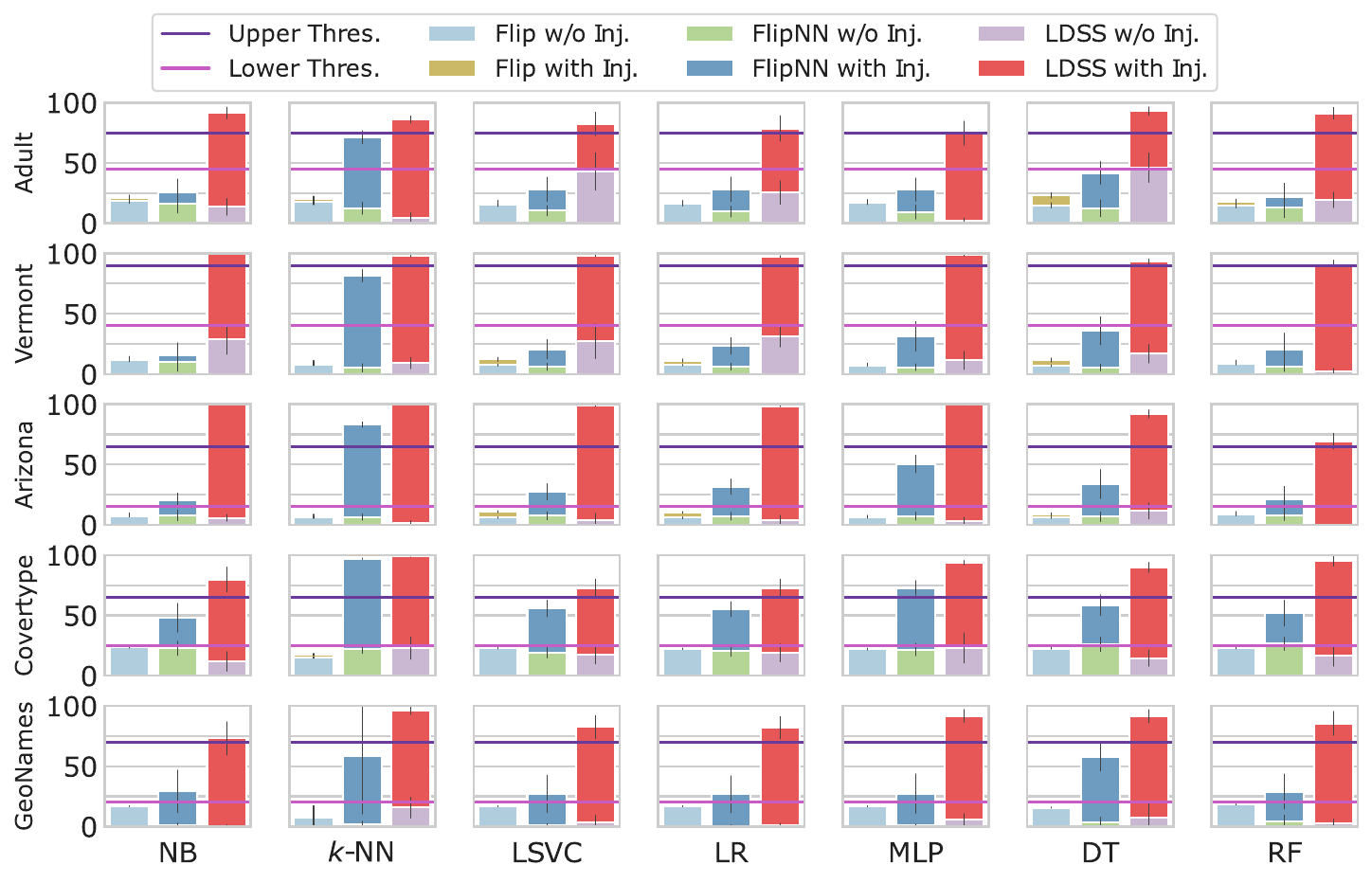}%
\caption{Trigger accuracy (\%) of classification models trained with and without $\mathcal{D}_{inj}$.}%
\label{fig:trigger_all}%
\end{figure}%

\subsection{Reliability and Robustness}
\label{sect:expt:rely_and_robust}

We assess whether a model is trained on $\mathcal{D}_{mod}$ by evaluating trigger accuracy against a pre-defined threshold. A model is likely trained on $\mathcal{D}_{mod}$ if its trigger accuracy surpasses this threshold, and vice versa.
A method is \emph{reliable} when there is a significant accuracy difference between models trained with and without $\mathcal{D}_{inj}$, and it is \emph{robust} if this difference remains large across various classification models.
Thus, we evaluate \textsc{LDSS}'s reliability and robustness by examining trigger accuracy across seven classification models, comparing it against two baselines, \textsc{Flip} and \textsc{FlipNN}.

The results are presented in Figure \ref{fig:trigger_all}, which stacks trigger accuracy of models trained on $\mathcal{D}_{orig}$ with and without $\mathcal{D}_{inj}$ on the same x-axis.
\textsc{LDSS} shows a consistent and pronounced difference in trigger accuracy for all models and datasets. In contrast, \textsc{Flip} and \textsc{FlipNN} lack this clear gap across most models.
The only exception is the good performance shown by \textsc{FlipNN} on the $k$-NN model as it flips a group of close samples locally to the same labels. 
Thus, it is much more likely to predict these triggers to the flipped label.

An ideal threshold shall be smaller than the trigger accuracy of any models trained on $\mathcal{D}_{mod}$ and larger than those trained on $\mathcal{D}_{orig}$. 
Following this principle, Figure \ref{fig:trigger_all} displays the upper and lower thresholds for \textsc{LDSS}, respectively denoted by dark and light purple lines.
This substantial threshold gap further highlights \textsc{LDSS}'s superior reliability and robustness over \textsc{Flip} and \textsc{FlipNN}.

\subsection{Fidelity}
\label{sect:expt:fidelity}

\begin{figure}[t]%
\centering%
\captionsetup{skip=0em,belowskip=0em}%
\includegraphics[width=0.95\columnwidth]{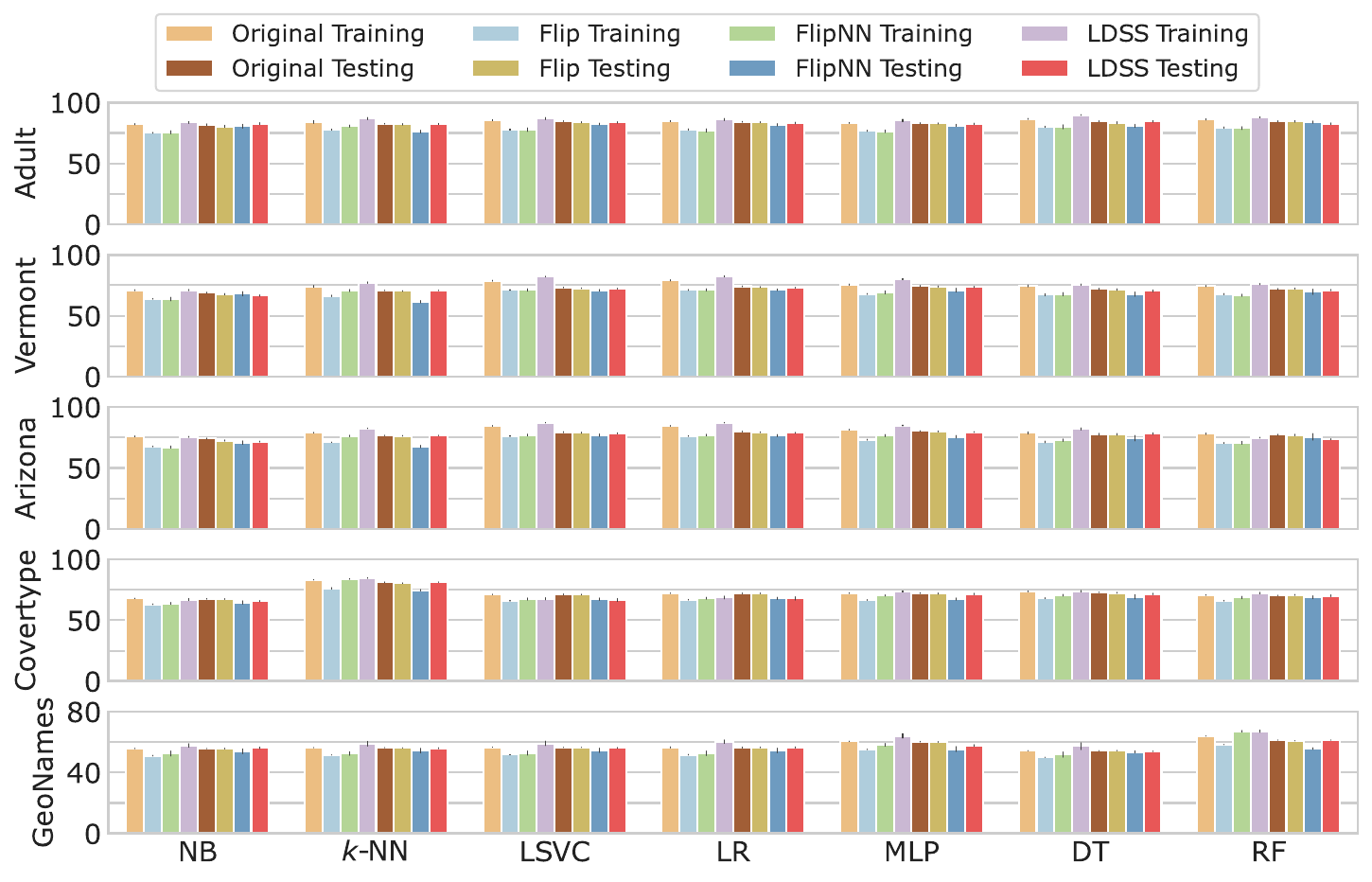}%
\caption{Training and testing accuracy (\%) of classification models trained on $\mathcal{D}_{orig}$ and $\mathcal{D}_{mod}$.}%
\label{fig:fidelity}%
\end{figure}

We examine \textsc{LDSS}'s fidelity by analyzing training and testing accuracy to verify a minimal impact on prediction accuracy by $\mathcal{D}_{mod}$.
Table \ref{tab:fidelity} shows that \textsc{LDSS} maintains low Average Absolute Difference (AAD) and Maximum Absolute Difference (MAD) across training and testing accuracies.
The largest MAD is a modest 4.57\% on Covertype, only a 6\% relative deviation from its 72.52\% testing accuracy. This confirms \textsc{LDSS}'s ability to preserve fidelity, reducing the likelihood of attackers detecting dataset changes.
Conversely, \textsc{Flip} and \textsc{FlipNN} show larger gaps in both AAD and MAD.
While \textsc{Flip}'s testing accuracy aligns with its random label flipping strategy, its training accuracy suffers due to misalignment with the model.
Figure \ref{fig:fidelity} further highlights that \textsc{LDSS} achieves minimal gaps in both training and testing accuracy across all models and datasets, evidenced by the small standard deviation in accuracy changes.
Compared to baseline methods, \textsc{LDSS} consistently has lower or comparable training and testing accuracy gaps for all models.

\begin{figure}[t]%
\centering%
\captionsetup{skip=0em,belowskip=0em}%
\subfigure[Outlier percentages (\%) under different contamination levels without dilution.]{%
  \label{fig:outlier_original}%
  \includegraphics[width=0.95\columnwidth]{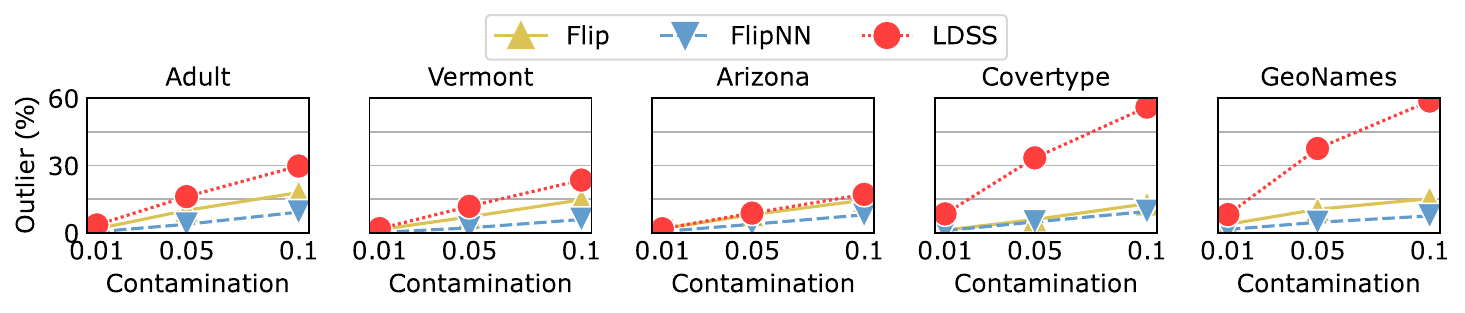}} \\%
\subfigure[\textsc{LDSS}'s outlier percentages (\%) under different dilution multipliers.]{%
  \label{fig:outlier_dilution}%
  \includegraphics[width=0.95\columnwidth]{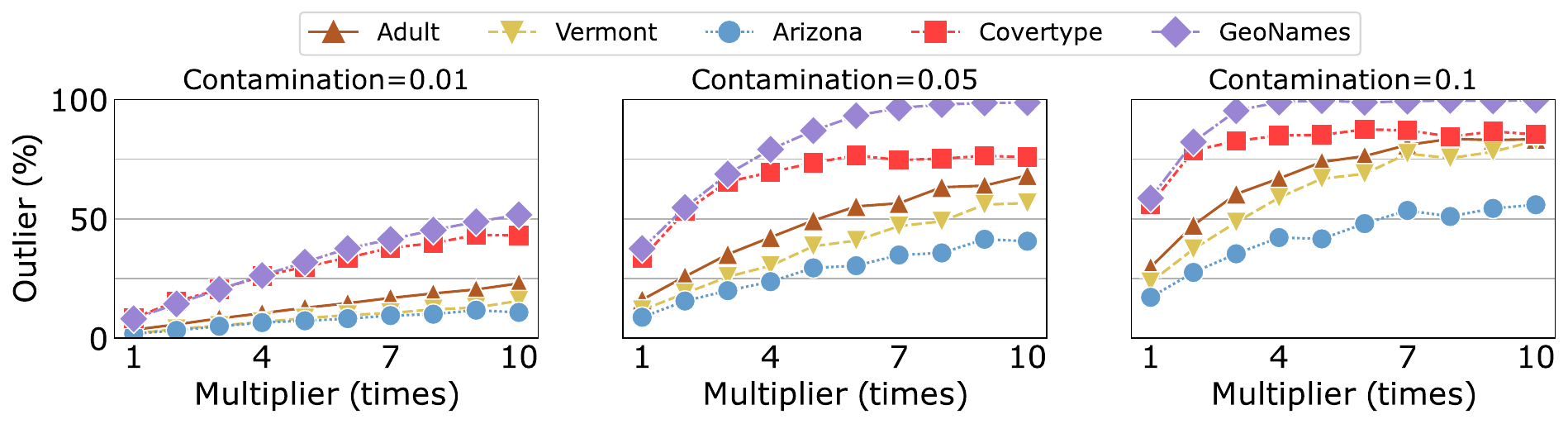}}%
\caption{Outlier percentages (\%) detected by Isolation Forest.}%
\label{fig:outlier}%
\end{figure}%

\begin{figure*}[t]%
\centering%
\captionsetup{skip=0.4em,belowskip=0em}%
\includegraphics[width=0.95\textwidth]{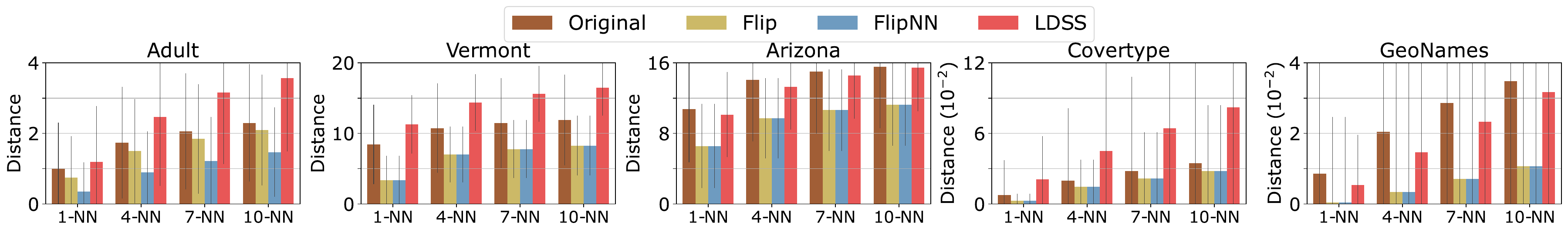}%
\caption{$1$-NN, $4$-NN, $7$-NN, and $10$-NN distances of original and injected samples by \textsc{Flip}, \textsc{FlipNN}, and \textsc{LDSS}.}%
\label{fig:dist_all}%
\end{figure*}%

\subsection{Security}
\label{sect:expt:security}

Under the assumption that attackers lack detailed knowledge of local distributions and marginal statistics, we assess \textsc{LDSS}'s security against unsupervised methods to identify injected samples.
We focus on whether \textsc{LDSS}'s injected samples are detectable via outlier and dense cluster detection and how resilient it is to dilution attacks, where attackers add more samples from similar populations.

\paragraph{Outlier Detection}
\textsc{LDSS} counters the risk of its injected samples $\mathcal{D}_{inj}$ being too distinct from the general population, which could make them vulnerable to outlier detection and easy removal by attackers. 
This is achieved by ensuring the centers of identified empty balls during the generation process are not excessively distant from existing data (refer to Section \ref{sect:method:empty_ball}). 
To validate this, we conduct outlier detection by Isolation Forest \cite{liu2008isolation} at contamination levels of 0.01, 0.05, and 0.1.
Figure \ref{fig:outlier_original} underlines the effectiveness of \textsc{LDSS}, despite a relatively higher outlier percentage (\%) compared to \textsc{Flip} and \textsc{FlipNN}. Even at the highest contamination level of 0.1, its values remain modest, peaking at around 60\% on Covertype and GeoNames.
Given a 10\% injection ratio ($\rho$), removing these outliers implies a concurrent loss of numerous genuine samples, adversely affecting the accuracy of models trained subsequently.

\paragraph{Dense Cluster Detection}
While injecting samples that form dense clusters can more effectively influence the model, they also increase detectability.
We assess this by analyzing the $k$-Nearest Neighbors distances ($k$-NN distances) within $\mathcal{D}_{mod}$, comparing those between original and injected samples.
As depicted in Figure \ref{fig:dist_all}, the $k$-NN distances of injected samples by \textsc{LDSS} are consistently similar or even larger than those of original datasets across various $k$ values. Thus, \textsc{LDSS} can evade dense cluster detection effectively.
In contrast, \textsc{Flip} and \textsc{FlipNN} introduce samples with significantly smaller $k$-NN distances, allowing attackers to accurately remove injected samples by searching for dense clusters.

\begin{figure}[t]%
\centering%
\captionsetup{skip=0em,belowskip=0em}%
\includegraphics[width=0.95\columnwidth]{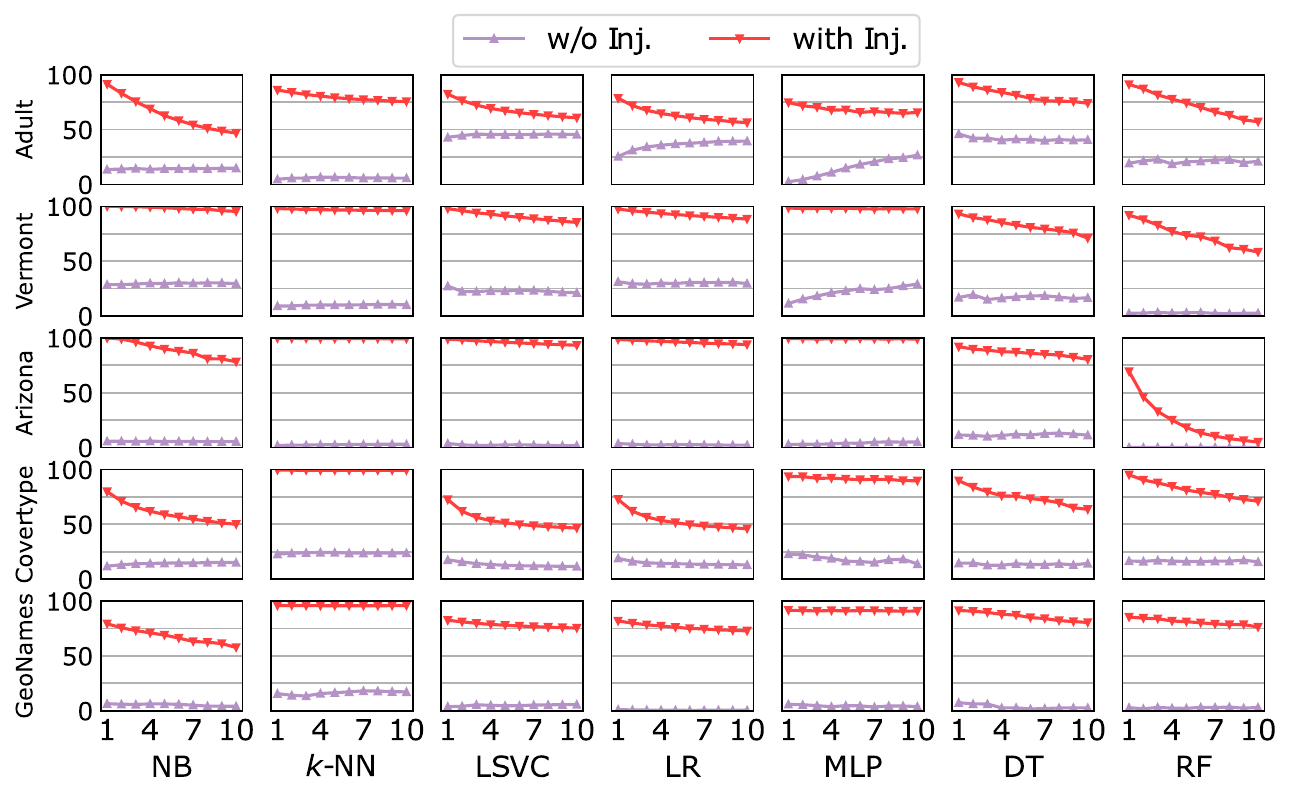}%
\caption{\textsc{LDSS}'s trigger accuracy (\%) of classification models under different dilution multipliers.}%
\label{fig:trigger_multiplier}%
\end{figure}

\begin{figure*}[t]%
\centering%
\captionsetup{skip=0em,belowskip=0em}%
\includegraphics[width=0.95\textwidth]{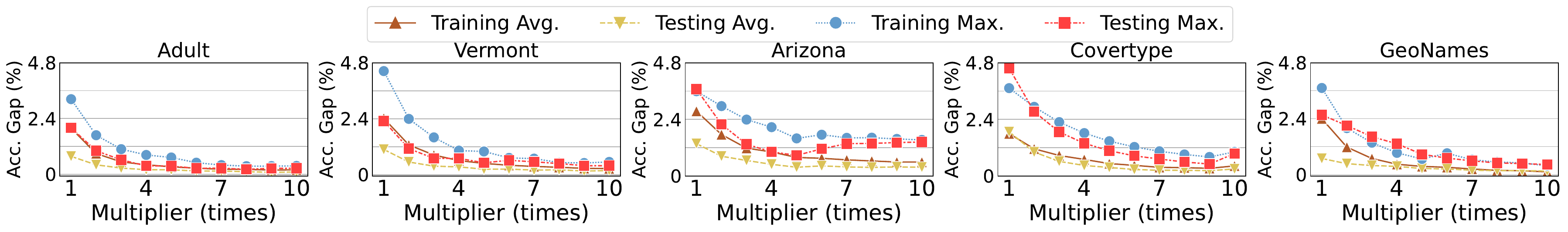}%
\caption{\textsc{LDSS}'s training and testing accuracy gap (\%) of classifications models under different dilution multipliers.}%
\label{fig:dil_fidelity}%
\end{figure*}

\begin{figure*}[t]%
\centering%
\captionsetup{skip=0em,belowskip=0em}%
\includegraphics[width=0.95\textwidth]{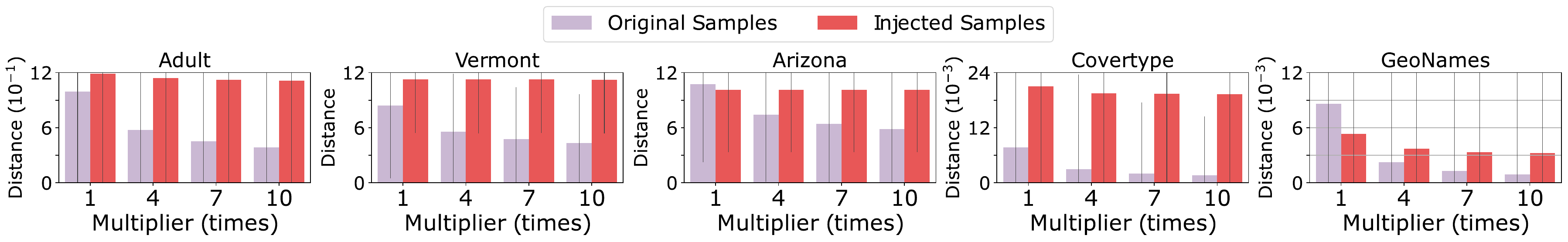}%
\caption{\textsc{LDSS}'s $1$-NN distance of original and injected samples under different dilution multipliers.}%
\label{fig:dil_dist}%
\end{figure*}

\paragraph{Dilution Attack}
We further measure whether \textsc{LDSS} is effective when attackers dilute the leaked dataset by incorporating data from other sources within the same population. 
Figures \ref{fig:outlier_dilution}, \ref{fig:trigger_multiplier}, \ref{fig:dil_fidelity}, and \ref{fig:dil_dist} collectively demonstrate \textsc{LDSS}'s ability to uphold high leaked data detection accuracy while preserving both fidelity and security.

Firstly, as evident from Figure \ref{fig:trigger_multiplier}, the gap in trigger accuracy remains substantial. Even under the dilution of 10 times, the accuracy gap remains large across almost all models and datasets, underscoring the resilience of \textsc{LDSS}.
Fidelity is expected to improve as the dataset becomes more diluted and injected samples comprise a smaller proportion--a conclusion affirmed by Figure \ref{fig:dil_fidelity}.
In terms of security, it is anticipated that a greater proportion of injected samples will be identifiable via outlier detection as the dataset becomes more diluted. 
As illustrated in Figure \ref{fig:outlier_dilution}, even under 10 times dilution, at most 50\% injected samples are flagged as outliers at contamination $=0.01$, and this number increases to 75\% and 85\% when contamination $= 0.05$ or $0.1$ among 4 out of 5 datasets.
In practice, this rarely happens as attackers need to gain access to multiple similar data sources; loss of model performance is expected by eliminating many original samples during outlier removal.
On the other hand, the likelihood of injected samples forming dense clusters decreases as dilution increases.
$k$-NN distances of \textsc{LDSS} injected samples are less apt to fall below those of genuine samples from both original and added datasets, as shown in Figure \ref{fig:dil_dist}.

\paragraph{Remarks}
Even in the scenario where attackers are aware of both $\mathcal{D}_{mod}$ and \textsc{LDSS}, the chances of a successful removal attack remain low, especially without extra data sources.
Firstly, pinpointing the exact empty balls used by \textsc{LDSS} for synthetic data injection is challenging as they are already filled and no longer empty. 
Moreover, re-running \textsc{LDSS} would probably yield different empty balls due to its heuristic nature of empty ball selection (see Section \ref{sect:method:synthetic}).
Attackers are then confined to using unsupervised methods to spot differences between injected and original samples. 
However, \textsc{LDSS} demonstrates strong defense capabilities against removal attacks, as injected samples exhibit similar $k$-NN distances to original ones (Figure \ref{fig:dist_all}) and are resilient to outlier detection (Figure \ref{fig:outlier_original}).

\subsection{Efficiency}
\label{sect:expt:efficiency}

We measure the efficiency through the total synthesis time.
At a 10\% injection ratio, \textsc{LDSS} completes within 2 hours for the largest GeoNames dataset--1.5 hours for empty ball identification and around 30 minutes for sample synthesis. 
For the smallest Adult dataset, this process takes only 30 minutes.
Considering this is a one-time process per dataset, the duration is notably brief.
However, extra time might be needed for extensive parameter tuning and performance evaluation, as training several classification models can be lengthy.
Optimization potential exists, particularly in accelerating the Nearest Neighbor Search (e.g., \cite{lei2019sublinear, lei2020locality, huang2021point}) in empty ball identification, where \textsc{LDSS} currently employs FAISS \cite{johnson2021billion}. 

\begin{figure}[t]%
\centering%
\captionsetup{skip=0em,belowskip=0em}%
\includegraphics[width=0.95\columnwidth]{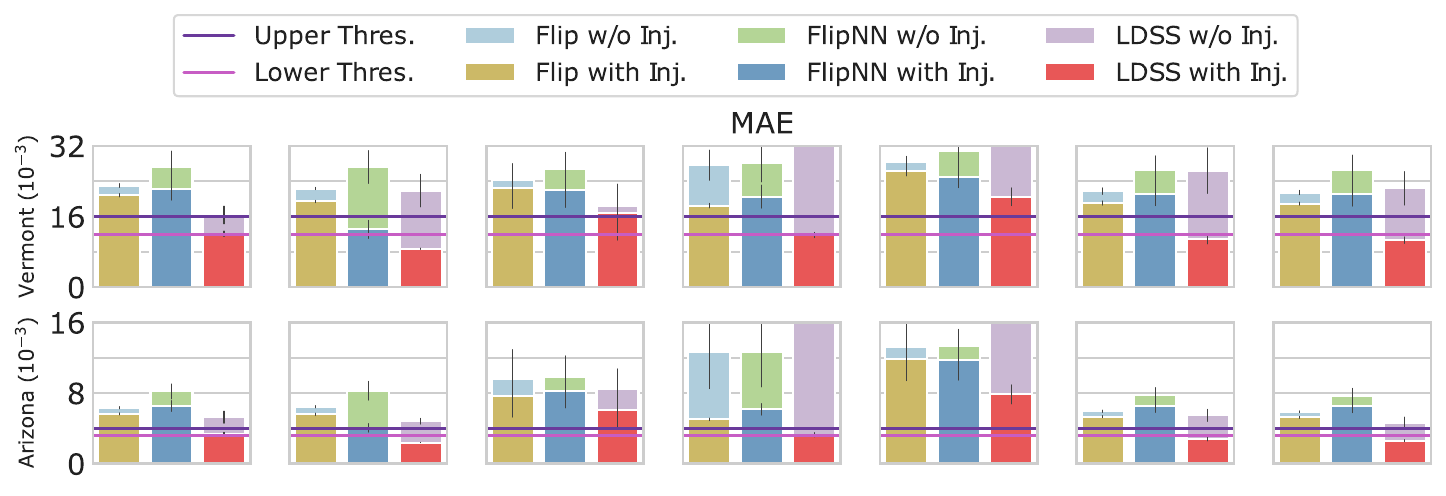} \\%
\includegraphics[width=0.95\columnwidth]{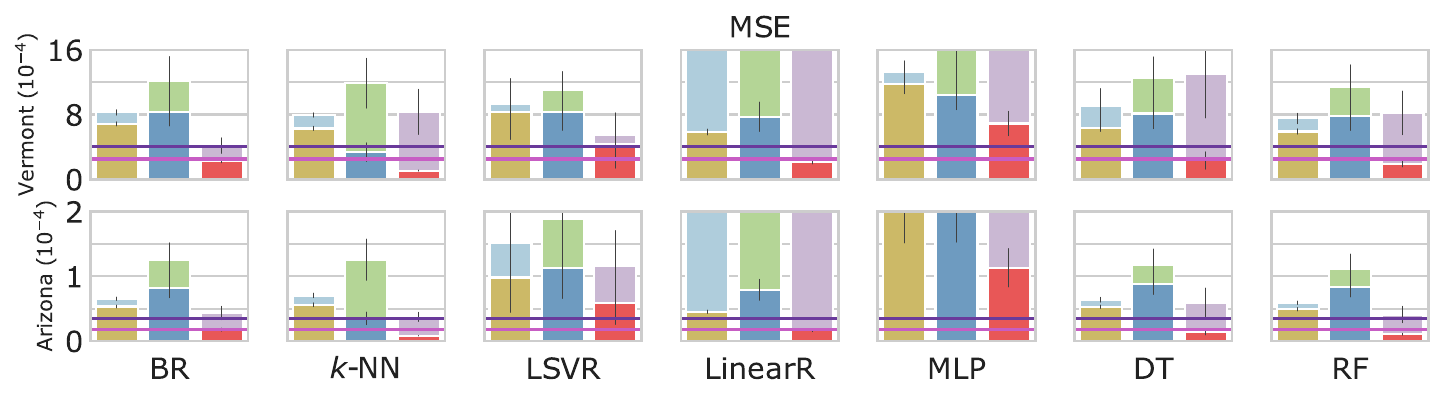} \\%
\caption{MAEs and MSEs for trigger samples predicted by regression models trained with and without $\mathcal{D}_{inj}$, assessed using the Vermont and Arizona datasets.}%
\label{fig:regression}%
\end{figure}%

\subsection{Regression}
\label{sect:expt:regression}

We extend \textsc{LDSS} to regression tasks to showcase its versatility, evaluating it with seven regression models: Bayes Ridge (BR), $k$-Nearest Neighbor ($k$-NN), Linear Support Vector Regression (LSVR), Linear Regression (LinearR), Multi-Layer Perceptron (MLP), Decision Tree (DT), and Random Forest (RF).
We use Mean Absolute Error (MAE) and Mean Squared Error (MSE) as evaluation measures, normalizing target variables to $[0,1]$. 
Arizona and Vermont are chosen for their rich numerical feature variety (e.g., INCWAGE) and their moderate size with a substantial number of features ideal for regression.

\begin{figure}[t]%
\centering%
\captionsetup{skip=0em,belowskip=0em}%
\includegraphics[width=0.95\columnwidth]{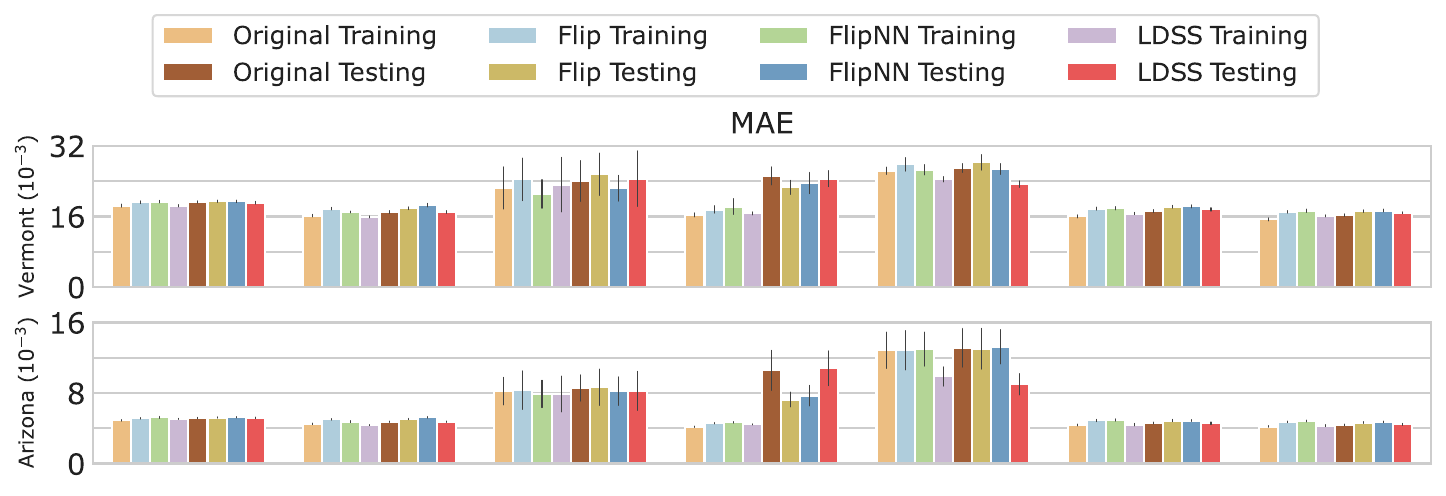} \\%
\includegraphics[width=0.95\columnwidth]{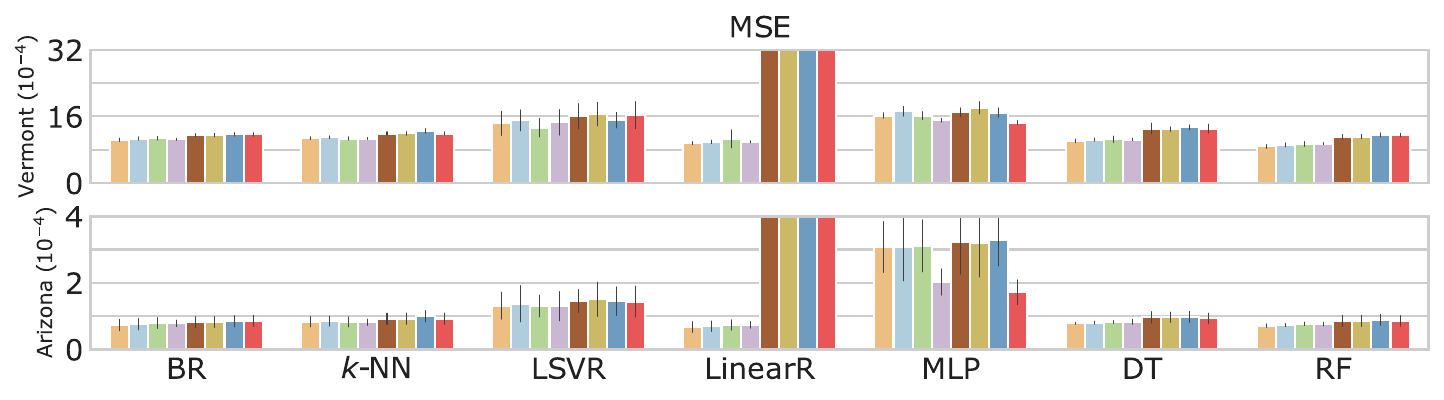} \\%
\caption{MAEs and MSEs for training and testing samples predicted by regression models trained on $\mathcal{D}_{orig}$ and $\mathcal{D}_{mod}$, assessed using the Vermont and Arizona datasets.}%
\label{fig:regression_fidelity}%
\end{figure}

Figure \ref{fig:regression} presents a similar layout to Figure \ref{fig:trigger_all}, where it stacks the MAE and MSE of trigger samples predicted by models trained with and without $\mathcal{D}_{inj}$ on the same x-axis.
Lower MAE or MSE suggests a model trained on $\mathcal{D}_{mod}$. As shown in Figure \ref{fig:trigger_all}, \textsc{LDSS} markedly reduces these metrics compared to \textsc{Flip} and \textsc{FlipNN}, demonstrating its efficacy in regression tasks.
Compared to classification tasks, the difference in MAEs and MSEs across models is smaller and more variable. This is mainly due to the higher variance of baseline MAEs and MSEs in regression than the accuracy variance in classification. 
In practice, it is advisable to choose a threshold that can minimize false alarms yet maintain detectability across most evaluated models, such as the upper and lower limits plotted in Figure \ref{fig:regression} based on \textsc{LDSS}'s results.
Despite being narrower, this gap still outperforms \textsc{Flip} and \textsc{FlipNN}, as \textsc{LDSS} generally yield lower MAEs and MSEs for trigger samples. 
Moreover, Figure \ref{fig:regression_fidelity} depicts that \textsc{LDSS} does not significantly affect the MAEs and MSEs of training and testing samples' prediction, ensuring good fidelity of models trained on the Vermont and Arizona datasets.

\subsection{Parameter Study}
\label{sect:expt:params}

\begin{figure}[t]%
\centering%
\captionsetup{skip=0em,belowskip=0em}%
\subfigure[Trigger Accuracy (\%).]{%
  \label{fig:para:trigger_gh}%
  \includegraphics[width=0.95\columnwidth]{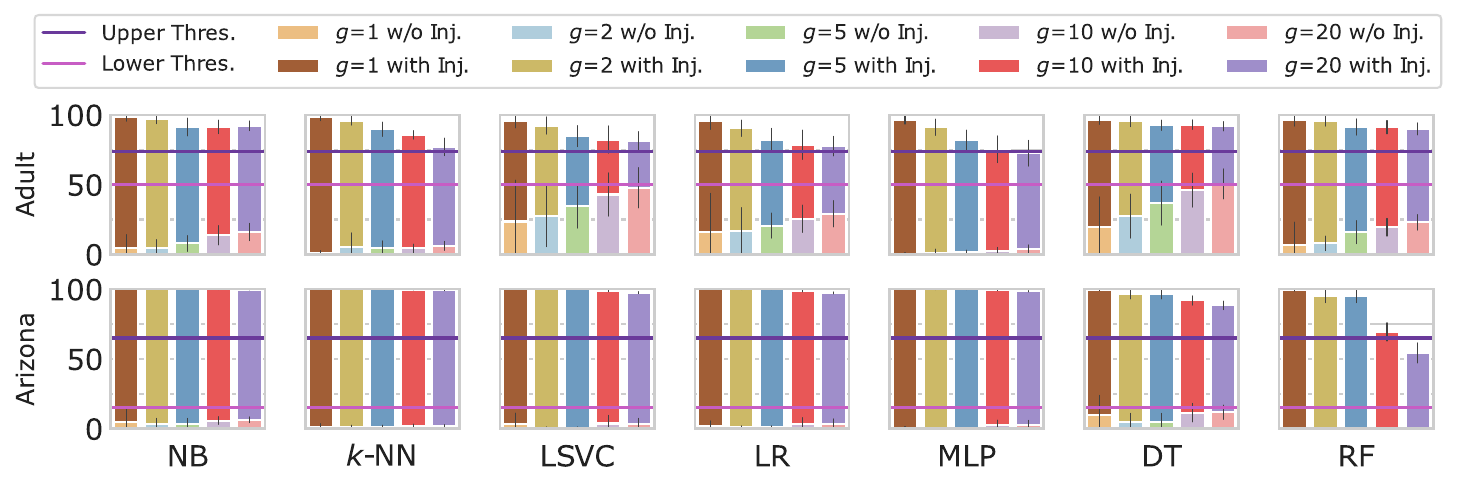}} \\%
\subfigure[$1$-NN Distance.]{%
  \label{fig:para:nndist_gh}%
  \includegraphics[width=0.95\columnwidth]{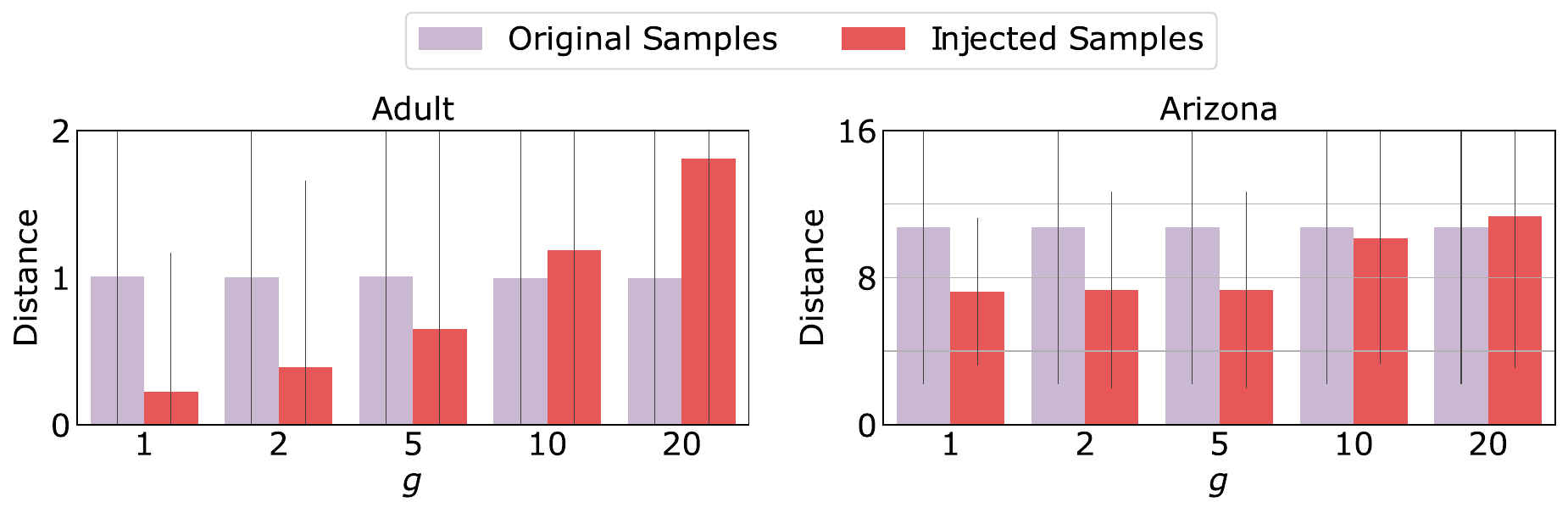}}%
\caption{\textsc{LDSS}'s trigger accuracy (\%) and $1$-NN distance under different $g$ values.}%
\label{fig:para}%
\end{figure}

\textsc{LDSS} hinges on two key parameters: $g$, the number of selected empty balls and $h$, the synthetic samples per ball.
Instead of direct specification, we calculate $h$ from an injection ratio $\rho$, i.e., $h=\rho n/g$, with $\rho$ fixed at 10\% for balanced reliability and security. 
We vary $g$ from 1 to 20 and present the results on Adult and Arizona in Figure \ref{fig:para}. Similar trends can be observed from other datasets.
Figure \ref{fig:para:trigger_gh} shows that a larger $g$ reduces the trigger accuracy gap due to fewer samples per ball, while Figure \ref{fig:para:nndist_gh} reveals that lower $g$ values (like 1, 2, and 5) result in smaller $1$-NN distances due to higher concentration in each ball.
An optimal balance is found at $g=10$, aligning NN distances with original data and ensuring a large enough trigger accuracy gap for effective detection.

\section{Related work}
\label{sect:related_work}

A plethora of research efforts have been devoted to safeguarding the IP of machine learning models through watermarking techniques \cite{li2021survey, xue2021intellectual, boenisch2021systematic, zhang2020model, zhang2021deep, fei2022supervised, peng2023intellectual, quan2020watermarking, wang2022protecting}. 
There are two broad categories of how watermark is embedded: either whinin training samples or directly into model parameters.
Additionally, membership inference attack \cite{shokri2017membership} provides a direct method to ascertain whether a model has been trained on specific samples.

\paragraph{Embedding Watermark in Training Samples}
Most methods employ a concept akin to model backdoors through data poisoning \cite{zhang2018protecting, adi2018turning, li2020open, chen2017targeted, li2020open, zhang2020model, namba2019robust, szyller2021dawn, wu2020watermarking, qiao2023novel, wang2022protecting, li2023plmmark}. 
The key concept is to train or finetune the target model with crafted samples by modifying existing or creating synthetic ones. 
The models are then trained to learn the specific pattern embedded with those crafted samples, either by mixing them into original training data or finetuning a trained model with only the crafted ones.
Recent research \cite{zhang2020model, yang2021robust, charette2022cosine, zhang2021exploring} aims to strengthen defense against potential watermark evasion or removal tactics. 
Predominantly, these approaches concentrate on DNNs, which, owing to their over-parameterization, are particularly adept at accommodating extensive information embedding.
In this work, we employ \textsc{Flip}, a basic label-flipping method, as our baseline,
and enhance it into \textsc{FlipNN} based on insights from methods like \cite{adi2018turning, namba2019robust}. We then develop \textsc{LDSS} to generate multiple samples within each empty ball, boosting watermark effectiveness.
\textsc{LDSS} achieves competitive detection quality to simpler models other than DNNs, ensuring that its injected samples are challenging to identify by closely mirroring the original dataset's distribution.
Unlike changing pixel patches in image tasks, \textsc{LDSS} targets the local distribution, a strategy that is more robust due to the typically lower local correlation of features in tabular data.

\paragraph{Embedding Watermark in Model Parameters}
Numerous studies concentrate on explicitly embedding watermark information into specific DNN layer parameters \cite{song2017machine, uchida2017embedding, wang2020watermarking}. 
During watermark verification or detection, they operate in a \emph{white-box} setting, requiring access to the model's parameters or gradient information. 
As such, detection involves retrieving parameters or activation of specific layers in response to given inputs and comparing them to the embedded watermarks after decoding.
On the other hand, in many real-world scenarios, model detection operates in a \emph{black-box} setting as models are usually accessed only as a service, returning predictions for given inputs. 
Thus, detecting watermarked models can only be performed by querying the target model with a specific trigger set to reveal the watermark in its predictions. 
This has led to advancements in model parameter embedding techniques for black-box detection \cite{chen2019blackmarks, darvish2019deepsigns, le2020adversarial}. 
Although they can provide higher accuracy and robustness than watermarking within training samples, they require complete control over the training process. 
Furthermore, they often necessitate white-box access to the model during detection. 
Our focus, however, is on safeguarding data IP without assumptions about the attackers' training approaches, making these methods inappropriate for our specific needs.

\paragraph{Membership Inference Attack}
Membership inference attack provides a direct approach to detecting leaked data \cite{shokri2017membership}.
This technique assesses the likelihood that samples from the owner's dataset have been used to train the target model, thereby identifying unauthorized usage. 
Nonetheless, this technique encounters challenges in discerning models trained on datasets with samples that are markedly similar or even identical to those in the owner's collection. 
For instance, a patient may have medical records at multiple hospitals, and these records could be substantially identical. 
Furthermore, its accuracy may diminish when applied to shallow models \cite{truex2021demystifying} like $k$-NN, Logistic Regression, or shallow Decision Trees and MLPs. 
Such models are less likely to exhibit significant differences in individual samples if they are trained on independent datasets drawn from the same population.

\section{Conclusion}
\label{sect:conclusion}

In this paper, we emphasize the criticality of protecting a dataset's IP, especially when it's leaked and misused. 
We introduce \textsc{LDSS}, an innovative, model-oblivious technique designed for detecting leaked data. 
By seamlessly incorporating synthesized samples into the dataset, \textsc{LDSS} adeptly determines if a model has been trained on the modified dataset. 
Through rigorous experiments, our results consistently affirm the reliability,
robustness, fidelity, security, and efficiency of \textsc{LDSS} in detecting leaked data across seven classification models. 
Its application in regression tasks further showcases its versatility and superior performance.

More importantly, we advocate for dataset IP protection, moving away from the conventional emphasis on just model IP. 
We have significantly mitigated the critical but often overlooked risks of model releases based on unauthorized datasets, marking a pivotal shift in machine learning IP security considerations. 
\textsc{LDSS} particularly resonates when access is restricted to the model as a service.

\begin{acks}
This research is supported by the National Research Foundation, Singapore under its AI Singapore Programme (AISG Award No: AISG3-RP-2022-029), the National Research Foundation, Singapore under its Strategic Capability Research Centres Funding Initiative, and the Ministry of Education, Singapore, under its MOE AcRF TIER 3 Grant (MOE-MOET32022-0001).
Any opinions, findings and conclusions or recommendations expressed in this material are those of the author(s) and do not reflect the views of National Research Foundation, Singapore and the Ministry of Education, Singapore.
\end{acks}


\balance
\bibliographystyle{ACM-Reference-Format}
\bibliography{main}


\begin{thebibliography}{55}


\ifx \showCODEN    \undefined \def \showCODEN     #1{\unskip}     \fi
\ifx \showDOI      \undefined \def \showDOI       #1{#1}\fi
\ifx \showISBNx    \undefined \def \showISBNx     #1{\unskip}     \fi
\ifx \showISBNxiii \undefined \def \showISBNxiii  #1{\unskip}     \fi
\ifx \showISSN     \undefined \def \showISSN      #1{\unskip}     \fi
\ifx \showLCCN     \undefined \def \showLCCN      #1{\unskip}     \fi
\ifx \shownote     \undefined \def \shownote      #1{#1}          \fi
\ifx \showarticletitle \undefined \def \showarticletitle #1{#1}   \fi
\ifx \showURL      \undefined \def \showURL       {\relax}        \fi
\providecommand\bibfield[2]{#2}
\providecommand\bibinfo[2]{#2}
\providecommand\natexlab[1]{#1}
\providecommand\showeprint[2][]{arXiv:#2}

\bibitem[\protect\citeauthoryear{Adi, Baum, Cisse, Pinkas, and Keshet}{Adi
  et~al\mbox{.}}{2018}]%
        {adi2018turning}
\bibfield{author}{\bibinfo{person}{Yossi Adi}, \bibinfo{person}{Carsten Baum},
  \bibinfo{person}{Moustapha Cisse}, \bibinfo{person}{Benny Pinkas}, {and}
  \bibinfo{person}{Joseph Keshet}.} \bibinfo{year}{2018}\natexlab{}.
\newblock \showarticletitle{Turning Your Weakness into a Strength: Watermarking
  Deep Neural Networks by Backdooring}. In \bibinfo{booktitle}{\emph{USENIX
  Security}}. \bibinfo{pages}{1615--1631}.
\newblock


\bibitem[\protect\citeauthoryear{Bellia}{Bellia}{2011}]%
        {bellia2011wikileaks}
\bibfield{author}{\bibinfo{person}{Patricia~L Bellia}.}
  \bibinfo{year}{2011}\natexlab{}.
\newblock \showarticletitle{WikiLeaks and the institutional framework for
  national security disclosures}.
\newblock \bibinfo{journal}{\emph{Yale LJ}}  \bibinfo{volume}{121}
  (\bibinfo{year}{2011}), \bibinfo{pages}{1448}.
\newblock


\bibitem[\protect\citeauthoryear{Bhatore, Mohan, and Reddy}{Bhatore
  et~al\mbox{.}}{2020}]%
        {bhatore2020machine}
\bibfield{author}{\bibinfo{person}{Siddharth Bhatore}, \bibinfo{person}{Lalit
  Mohan}, {and} \bibinfo{person}{Y~Raghu Reddy}.}
  \bibinfo{year}{2020}\natexlab{}.
\newblock \showarticletitle{Machine learning techniques for credit risk
  evaluation: a systematic literature review}.
\newblock \bibinfo{journal}{\emph{Journal of Banking and Financial Technology}}
   \bibinfo{volume}{4} (\bibinfo{year}{2020}), \bibinfo{pages}{111--138}.
\newblock


\bibitem[\protect\citeauthoryear{Bi, Sun, Huang, Yang, and Huang}{Bi
  et~al\mbox{.}}{2007}]%
        {bi2007robust}
\bibfield{author}{\bibinfo{person}{Ning Bi}, \bibinfo{person}{Qiyu Sun},
  \bibinfo{person}{Daren Huang}, \bibinfo{person}{Zhihua Yang}, {and}
  \bibinfo{person}{Jiwu Huang}.} \bibinfo{year}{2007}\natexlab{}.
\newblock \showarticletitle{Robust Image Watermarking based on Multiband
  Wavelets and Empirical Mode Decomposition}.
\newblock \bibinfo{journal}{\emph{TIP}} \bibinfo{volume}{16},
  \bibinfo{number}{8} (\bibinfo{year}{2007}), \bibinfo{pages}{1956--1966}.
\newblock


\bibitem[\protect\citeauthoryear{Boenisch}{Boenisch}{2021}]%
        {boenisch2021systematic}
\bibfield{author}{\bibinfo{person}{Franziska Boenisch}.}
  \bibinfo{year}{2021}\natexlab{}.
\newblock \showarticletitle{A Systematic Review on Model Watermarking for
  Neural Networks}.
\newblock \bibinfo{journal}{\emph{Frontiers in Big Data}}  \bibinfo{volume}{4}
  (\bibinfo{year}{2021}), \bibinfo{pages}{729663}.
\newblock


\bibitem[\protect\citeauthoryear{Charette, Chu, Chen, Pei, Wang, and
  Zhang}{Charette et~al\mbox{.}}{2022}]%
        {charette2022cosine}
\bibfield{author}{\bibinfo{person}{Laurent Charette}, \bibinfo{person}{Lingyang
  Chu}, \bibinfo{person}{Yizhou Chen}, \bibinfo{person}{Jian Pei},
  \bibinfo{person}{Lanjun Wang}, {and} \bibinfo{person}{Yong Zhang}.}
  \bibinfo{year}{2022}\natexlab{}.
\newblock \showarticletitle{Cosine model watermarking against ensemble
  distillation}. In \bibinfo{booktitle}{\emph{AAAI}},
  Vol.~\bibinfo{volume}{36}. \bibinfo{pages}{9512--9520}.
\newblock


\bibitem[\protect\citeauthoryear{Chawla, Bowyer, Hall, and Kegelmeyer}{Chawla
  et~al\mbox{.}}{2002}]%
        {chawla2002smote}
\bibfield{author}{\bibinfo{person}{Nitesh~V Chawla}, \bibinfo{person}{Kevin~W
  Bowyer}, \bibinfo{person}{Lawrence~O Hall}, {and} \bibinfo{person}{W~Philip
  Kegelmeyer}.} \bibinfo{year}{2002}\natexlab{}.
\newblock \showarticletitle{SMOTE: Synthetic Minority Over-Sampling Technique}.
\newblock \bibinfo{journal}{\emph{JAIR}}  \bibinfo{volume}{16}
  (\bibinfo{year}{2002}), \bibinfo{pages}{321--357}.
\newblock


\bibitem[\protect\citeauthoryear{Chen, Rouhani, and Koushanfar}{Chen
  et~al\mbox{.}}{2019}]%
        {chen2019blackmarks}
\bibfield{author}{\bibinfo{person}{Huili Chen}, \bibinfo{person}{Bita~Darvish
  Rouhani}, {and} \bibinfo{person}{Farinaz Koushanfar}.}
  \bibinfo{year}{2019}\natexlab{}.
\newblock \showarticletitle{Blackmarks: Blackbox Multibit Watermarking for Deep
  Neural Networks}.
\newblock \bibinfo{journal}{\emph{arXiv preprint arXiv:1904.00344}}
  (\bibinfo{year}{2019}).
\newblock


\bibitem[\protect\citeauthoryear{Chen, Liu, Li, Lu, and Song}{Chen
  et~al\mbox{.}}{2017}]%
        {chen2017targeted}
\bibfield{author}{\bibinfo{person}{Xinyun Chen}, \bibinfo{person}{Chang Liu},
  \bibinfo{person}{Bo Li}, \bibinfo{person}{Kimberly Lu}, {and}
  \bibinfo{person}{Dawn Song}.} \bibinfo{year}{2017}\natexlab{}.
\newblock \showarticletitle{Targeted Backdoor Attacks on Deep Learning Systems
  using Data Poisoning}.
\newblock \bibinfo{journal}{\emph{arXiv preprint arXiv:1712.05526}}
  (\bibinfo{year}{2017}).
\newblock


\bibitem[\protect\citeauthoryear{Cheng, Liu, and Yao}{Cheng
  et~al\mbox{.}}{2017}]%
        {cheng2017enterprise}
\bibfield{author}{\bibinfo{person}{Long Cheng}, \bibinfo{person}{Fang Liu},
  {and} \bibinfo{person}{Danfeng Yao}.} \bibinfo{year}{2017}\natexlab{}.
\newblock \showarticletitle{Enterprise data breach: causes, challenges,
  prevention, and future directions}.
\newblock \bibinfo{journal}{\emph{Wiley Interdisciplinary Reviews: Data Mining
  and Knowledge Discovery}} \bibinfo{volume}{7}, \bibinfo{number}{5}
  (\bibinfo{year}{2017}), \bibinfo{pages}{e1211}.
\newblock


\bibitem[\protect\citeauthoryear{Cox, Miller, Bloom, Fridrich, and Kalker}{Cox
  et~al\mbox{.}}{2007}]%
        {cox2007digital}
\bibfield{author}{\bibinfo{person}{Ingemar Cox}, \bibinfo{person}{Matthew
  Miller}, \bibinfo{person}{Jeffrey Bloom}, \bibinfo{person}{Jessica Fridrich},
  {and} \bibinfo{person}{Ton Kalker}.} \bibinfo{year}{2007}\natexlab{}.
\newblock \bibinfo{booktitle}{\emph{Digital Watermarking and Steganography}}.
\newblock \bibinfo{publisher}{Morgan Kaufmann}.
\newblock


\bibitem[\protect\citeauthoryear{Darvish~Rouhani, Chen, and
  Koushanfar}{Darvish~Rouhani et~al\mbox{.}}{2019}]%
        {darvish2019deepsigns}
\bibfield{author}{\bibinfo{person}{Bita Darvish~Rouhani},
  \bibinfo{person}{Huili Chen}, {and} \bibinfo{person}{Farinaz Koushanfar}.}
  \bibinfo{year}{2019}\natexlab{}.
\newblock \showarticletitle{Deepsigns: An End-to-End Watermarking Framework for
  Ownership Protection of Deep Neural Networks}. In
  \bibinfo{booktitle}{\emph{ASPLOS}}. \bibinfo{pages}{485--497}.
\newblock


\bibitem[\protect\citeauthoryear{Dua and Graff}{Dua and Graff}{2017}]%
        {Dua:2019}
\bibfield{author}{\bibinfo{person}{Dheeru Dua} {and} \bibinfo{person}{Casey
  Graff}.} \bibinfo{year}{2017}\natexlab{}.
\newblock \bibinfo{title}{{UCI} Machine Learning Repository}.
\newblock
\newblock
\urldef\tempurl%
\url{http://archive.ics.uci.edu/ml}
\showURL{%
\tempurl}


\bibitem[\protect\citeauthoryear{Fei, Xia, Tondi, and Barni}{Fei
  et~al\mbox{.}}{2022}]%
        {fei2022supervised}
\bibfield{author}{\bibinfo{person}{Jianwei Fei}, \bibinfo{person}{Zhihua Xia},
  \bibinfo{person}{Benedetta Tondi}, {and} \bibinfo{person}{Mauro Barni}.}
  \bibinfo{year}{2022}\natexlab{}.
\newblock \showarticletitle{Supervised {GAN} Watermarking for Intellectual
  Property Protection}. In \bibinfo{booktitle}{\emph{IEEE International
  Workshop on Information Forensics and Security}}. \bibinfo{pages}{1--6}.
\newblock


\bibitem[\protect\citeauthoryear{Halko, Martinsson, and Tropp}{Halko
  et~al\mbox{.}}{2011}]%
        {halko2011finding}
\bibfield{author}{\bibinfo{person}{Nathan Halko}, \bibinfo{person}{Per-Gunnar
  Martinsson}, {and} \bibinfo{person}{Joel~A Tropp}.}
  \bibinfo{year}{2011}\natexlab{}.
\newblock \showarticletitle{Finding Structure with Randomness: Probabilistic
  Algorithms for Constructing Approximate Matrix Decompositions}.
\newblock \bibinfo{journal}{\emph{SIAM Rev.}} \bibinfo{volume}{53},
  \bibinfo{number}{2} (\bibinfo{year}{2011}), \bibinfo{pages}{217--288}.
\newblock


\bibitem[\protect\citeauthoryear{Huang, Shi, and Shi}{Huang
  et~al\mbox{.}}{2000}]%
        {huang2000embedding}
\bibfield{author}{\bibinfo{person}{Jiwu Huang}, \bibinfo{person}{Yun~Q Shi},
  {and} \bibinfo{person}{Yi Shi}.} \bibinfo{year}{2000}\natexlab{}.
\newblock \showarticletitle{Embedding Image Watermarks in DC Components}.
\newblock \bibinfo{journal}{\emph{TCSVT}} \bibinfo{volume}{10},
  \bibinfo{number}{6} (\bibinfo{year}{2000}), \bibinfo{pages}{974--979}.
\newblock


\bibitem[\protect\citeauthoryear{Huang, Lei, and Tung}{Huang
  et~al\mbox{.}}{2021}]%
        {huang2021point}
\bibfield{author}{\bibinfo{person}{Qiang Huang}, \bibinfo{person}{Yifan Lei},
  {and} \bibinfo{person}{Anthony~KH Tung}.} \bibinfo{year}{2021}\natexlab{}.
\newblock \showarticletitle{Point-to-Hyperplane Nearest Neighbor Search Beyond
  the Unit Hypersphere}. In \bibinfo{booktitle}{\emph{SIGMOD}}.
  \bibinfo{pages}{777--789}.
\newblock


\bibitem[\protect\citeauthoryear{Huang, McCullagh, Black, and Harper}{Huang
  et~al\mbox{.}}{2007}]%
        {huang2007feature}
\bibfield{author}{\bibinfo{person}{Yue Huang}, \bibinfo{person}{Paul
  McCullagh}, \bibinfo{person}{Norman Black}, {and} \bibinfo{person}{Roy
  Harper}.} \bibinfo{year}{2007}\natexlab{}.
\newblock \showarticletitle{Feature selection and classification model
  construction on type 2 diabetic patients’ data}.
\newblock \bibinfo{journal}{\emph{Artificial Intelligence in Medicine}}
  \bibinfo{volume}{41}, \bibinfo{number}{3} (\bibinfo{year}{2007}),
  \bibinfo{pages}{251--262}.
\newblock


\bibitem[\protect\citeauthoryear{Johnson, Douze, and Jegou}{Johnson
  et~al\mbox{.}}{2021}]%
        {johnson2021billion}
\bibfield{author}{\bibinfo{person}{Jeff Johnson}, \bibinfo{person}{Matthijs
  Douze}, {and} \bibinfo{person}{Herve Jegou}.}
  \bibinfo{year}{2021}\natexlab{}.
\newblock \showarticletitle{Billion-Scale Similarity Search with GPUs}.
\newblock \bibinfo{journal}{\emph{TBD}} \bibinfo{volume}{7},
  \bibinfo{number}{3} (\bibinfo{year}{2021}), \bibinfo{pages}{535--547}.
\newblock


\bibitem[\protect\citeauthoryear{Knuth}{Knuth}{2014}]%
        {knuth2014art}
\bibfield{author}{\bibinfo{person}{Donald~E Knuth}.}
  \bibinfo{year}{2014}\natexlab{}.
\newblock \bibinfo{booktitle}{\emph{The Art of Computer Programming:
  Seminumerical Algorithms, volume 2}}.
\newblock \bibinfo{publisher}{Addison-Wesley Professional}.
\newblock


\bibitem[\protect\citeauthoryear{Le~Merrer, Perez, and Tr{\'e}dan}{Le~Merrer
  et~al\mbox{.}}{2020}]%
        {le2020adversarial}
\bibfield{author}{\bibinfo{person}{Erwan Le~Merrer}, \bibinfo{person}{Patrick
  Perez}, {and} \bibinfo{person}{Gilles Tr{\'e}dan}.}
  \bibinfo{year}{2020}\natexlab{}.
\newblock \showarticletitle{Adversarial Frontier Stitching for Remote Neural
  Network Watermarking}.
\newblock \bibinfo{journal}{\emph{Neural Computing and Applications}}
  \bibinfo{volume}{32}, \bibinfo{number}{13} (\bibinfo{year}{2020}),
  \bibinfo{pages}{9233--9244}.
\newblock


\bibitem[\protect\citeauthoryear{Lee, Cho, Lee, Jang, Jang, Nam, and Park}{Lee
  et~al\mbox{.}}{2004}]%
        {lee2004stochastic}
\bibfield{author}{\bibinfo{person}{Jong-Seok Lee}, \bibinfo{person}{Taeg-Sang
  Cho}, \bibinfo{person}{Jiye Lee}, \bibinfo{person}{Myung-Kee Jang},
  \bibinfo{person}{Tae-Kwang Jang}, \bibinfo{person}{Dongkyung Nam}, {and}
  \bibinfo{person}{Cheol~Hoon Park}.} \bibinfo{year}{2004}\natexlab{}.
\newblock \showarticletitle{A Stochastic Search Approach for the
  Multidimensional Largest Empty Sphere Problem}.
\newblock  (\bibinfo{year}{2004}), \bibinfo{pages}{1--11}.
\newblock


\bibitem[\protect\citeauthoryear{Lei, Huang, Kankanhalli, and Tung}{Lei
  et~al\mbox{.}}{2019}]%
        {lei2019sublinear}
\bibfield{author}{\bibinfo{person}{Yifan Lei}, \bibinfo{person}{Qiang Huang},
  \bibinfo{person}{Mohan Kankanhalli}, {and} \bibinfo{person}{Anthony Tung}.}
  \bibinfo{year}{2019}\natexlab{}.
\newblock \showarticletitle{Sublinear Time Nearest Neighbor Search over
  Generalized Weighted Space}. In \bibinfo{booktitle}{\emph{ICML}}.
  \bibinfo{pages}{3773--3781}.
\newblock


\bibitem[\protect\citeauthoryear{Lei, Huang, Kankanhalli, and Tung}{Lei
  et~al\mbox{.}}{2020}]%
        {lei2020locality}
\bibfield{author}{\bibinfo{person}{Yifan Lei}, \bibinfo{person}{Qiang Huang},
  \bibinfo{person}{Mohan Kankanhalli}, {and} \bibinfo{person}{Anthony~KH
  Tung}.} \bibinfo{year}{2020}\natexlab{}.
\newblock \showarticletitle{Locality-Sensitive Hashing Scheme based on Longest
  Circular Co-Substring}. In \bibinfo{booktitle}{\emph{SIGMOD}}.
  \bibinfo{pages}{2589--2599}.
\newblock


\bibitem[\protect\citeauthoryear{Li, Cheng, Li, Du, Zhao, and Liu}{Li
  et~al\mbox{.}}{2023}]%
        {li2023plmmark}
\bibfield{author}{\bibinfo{person}{Peixuan Li}, \bibinfo{person}{Pengzhou
  Cheng}, \bibinfo{person}{Fangqi Li}, \bibinfo{person}{Wei Du},
  \bibinfo{person}{Haodong Zhao}, {and} \bibinfo{person}{Gongshen Liu}.}
  \bibinfo{year}{2023}\natexlab{}.
\newblock \showarticletitle{PLMmark: a secure and robust black-box watermarking
  framework for pre-trained language models}. In
  \bibinfo{booktitle}{\emph{AAAI}}, Vol.~\bibinfo{volume}{37}.
  \bibinfo{pages}{14991--14999}.
\newblock


\bibitem[\protect\citeauthoryear{Li, Wang, and Barni}{Li et~al\mbox{.}}{2021}]%
        {li2021survey}
\bibfield{author}{\bibinfo{person}{Yue Li}, \bibinfo{person}{Hongxia Wang},
  {and} \bibinfo{person}{Mauro Barni}.} \bibinfo{year}{2021}\natexlab{}.
\newblock \showarticletitle{A Survey of Deep Neural Network Watermarking
  Techniques}.
\newblock \bibinfo{journal}{\emph{Neurocomputing}}  \bibinfo{volume}{461}
  (\bibinfo{year}{2021}), \bibinfo{pages}{171--193}.
\newblock


\bibitem[\protect\citeauthoryear{Li, Zhang, Bai, Wu, Jiang, and Xia}{Li
  et~al\mbox{.}}{2020}]%
        {li2020open}
\bibfield{author}{\bibinfo{person}{Yiming Li}, \bibinfo{person}{Ziqi Zhang},
  \bibinfo{person}{Jiawang Bai}, \bibinfo{person}{Baoyuan Wu},
  \bibinfo{person}{Yong Jiang}, {and} \bibinfo{person}{Shu-Tao Xia}.}
  \bibinfo{year}{2020}\natexlab{}.
\newblock \showarticletitle{Open-sourced Dataset Protection via Backdoor
  Watermarking}.
\newblock \bibinfo{journal}{\emph{arXiv preprint arXiv:2010.05821}}
  (\bibinfo{year}{2020}).
\newblock


\bibitem[\protect\citeauthoryear{Liu, Ting, and Zhou}{Liu
  et~al\mbox{.}}{2008}]%
        {liu2008isolation}
\bibfield{author}{\bibinfo{person}{Fei~Tony Liu}, \bibinfo{person}{Kai~Ming
  Ting}, {and} \bibinfo{person}{Zhi-Hua Zhou}.}
  \bibinfo{year}{2008}\natexlab{}.
\newblock \showarticletitle{Isolation Forest}. In
  \bibinfo{booktitle}{\emph{ICDM}}. \bibinfo{pages}{413--422}.
\newblock


\bibitem[\protect\citeauthoryear{Minka}{Minka}{2000}]%
        {minka2000automatic}
\bibfield{author}{\bibinfo{person}{Thomas~P Minka}.}
  \bibinfo{year}{2000}\natexlab{}.
\newblock \showarticletitle{Automatic Choice of Dimensionality for PCA}. In
  \bibinfo{booktitle}{\emph{NIPS}}. \bibinfo{pages}{577--583}.
\newblock


\bibitem[\protect\citeauthoryear{Namba and Sakuma}{Namba and Sakuma}{2019}]%
        {namba2019robust}
\bibfield{author}{\bibinfo{person}{Ryota Namba} {and} \bibinfo{person}{Jun
  Sakuma}.} \bibinfo{year}{2019}\natexlab{}.
\newblock \showarticletitle{Robust Watermarking of Neural Network with
  Exponential Weighting}. In \bibinfo{booktitle}{\emph{AsiaCCS}}.
  \bibinfo{pages}{228--240}.
\newblock


\bibitem[\protect\citeauthoryear{Palaniappan and Mandic}{Palaniappan and
  Mandic}{2007}]%
        {palaniappan2007biometrics}
\bibfield{author}{\bibinfo{person}{Ramaswamy Palaniappan} {and}
  \bibinfo{person}{Danilo~P Mandic}.} \bibinfo{year}{2007}\natexlab{}.
\newblock \showarticletitle{Biometrics from brain electrical activity: A
  machine learning approach}.
\newblock \bibinfo{journal}{\emph{TPAMI}} \bibinfo{volume}{29},
  \bibinfo{number}{4} (\bibinfo{year}{2007}), \bibinfo{pages}{738--742}.
\newblock


\bibitem[\protect\citeauthoryear{Peng, Chen, Xu, Chen, Wang, and Jia}{Peng
  et~al\mbox{.}}{2023}]%
        {peng2023intellectual}
\bibfield{author}{\bibinfo{person}{Sen Peng}, \bibinfo{person}{Yufei Chen},
  \bibinfo{person}{Jie Xu}, \bibinfo{person}{Zizhuo Chen},
  \bibinfo{person}{Cong Wang}, {and} \bibinfo{person}{Xiaohua Jia}.}
  \bibinfo{year}{2023}\natexlab{}.
\newblock \showarticletitle{Intellectual Property Protection of DNN Models}.
\newblock \bibinfo{journal}{\emph{World Wide Web}} \bibinfo{volume}{26},
  \bibinfo{number}{4} (\bibinfo{year}{2023}), \bibinfo{pages}{1877--1911}.
\newblock


\bibitem[\protect\citeauthoryear{Potdar, Han, and Chang}{Potdar
  et~al\mbox{.}}{2005}]%
        {potdar2005survey}
\bibfield{author}{\bibinfo{person}{Vidyasagar~M Potdar}, \bibinfo{person}{Song
  Han}, {and} \bibinfo{person}{Elizabeth Chang}.}
  \bibinfo{year}{2005}\natexlab{}.
\newblock \showarticletitle{A Survey of Digital Image Watermarking Techniques}.
  In \bibinfo{booktitle}{\emph{IEEE International Conference on Industrial
  Informatics}}. \bibinfo{pages}{709--716}.
\newblock


\bibitem[\protect\citeauthoryear{Qiao, Ma, Zheng, Wu, Chen, Xu, and Luo}{Qiao
  et~al\mbox{.}}{2023}]%
        {qiao2023novel}
\bibfield{author}{\bibinfo{person}{Tong Qiao}, \bibinfo{person}{Yuyan Ma},
  \bibinfo{person}{Ning Zheng}, \bibinfo{person}{Hanzhou Wu},
  \bibinfo{person}{Yanli Chen}, \bibinfo{person}{Ming Xu}, {and}
  \bibinfo{person}{Xiangyang Luo}.} \bibinfo{year}{2023}\natexlab{}.
\newblock \showarticletitle{A novel model watermarking for protecting
  generative adversarial network}.
\newblock \bibinfo{journal}{\emph{Computers \& Security}}
  \bibinfo{volume}{127} (\bibinfo{year}{2023}), \bibinfo{pages}{103102}.
\newblock


\bibitem[\protect\citeauthoryear{Quan, Teng, Chen, and Ji}{Quan
  et~al\mbox{.}}{2020}]%
        {quan2020watermarking}
\bibfield{author}{\bibinfo{person}{Yuhui Quan}, \bibinfo{person}{Huan Teng},
  \bibinfo{person}{Yixin Chen}, {and} \bibinfo{person}{Hui Ji}.}
  \bibinfo{year}{2020}\natexlab{}.
\newblock \showarticletitle{Watermarking deep neural networks in image
  processing}.
\newblock \bibinfo{journal}{\emph{TNNLS}} \bibinfo{volume}{32},
  \bibinfo{number}{5} (\bibinfo{year}{2020}), \bibinfo{pages}{1852--1865}.
\newblock


\bibitem[\protect\citeauthoryear{Ribeiro, Grolinger, and Capretz}{Ribeiro
  et~al\mbox{.}}{2015}]%
        {ribeiro2015mlaas}
\bibfield{author}{\bibinfo{person}{Mauro Ribeiro}, \bibinfo{person}{Katarina
  Grolinger}, {and} \bibinfo{person}{Miriam~AM Capretz}.}
  \bibinfo{year}{2015}\natexlab{}.
\newblock \showarticletitle{Mlaas: Machine learning as a service}. In
  \bibinfo{booktitle}{\emph{IEEE 14th International Conference on Machine
  Learning and Applications}}. \bibinfo{pages}{896--902}.
\newblock


\bibitem[\protect\citeauthoryear{Saha, Subramanya, and Pirsiavash}{Saha
  et~al\mbox{.}}{2020}]%
        {saha2020hidden}
\bibfield{author}{\bibinfo{person}{Aniruddha Saha},
  \bibinfo{person}{Akshayvarun Subramanya}, {and} \bibinfo{person}{Hamed
  Pirsiavash}.} \bibinfo{year}{2020}\natexlab{}.
\newblock \showarticletitle{Hidden trigger backdoor attacks}. In
  \bibinfo{booktitle}{\emph{AAAI}}, Vol.~\bibinfo{volume}{34}.
  \bibinfo{pages}{11957--11965}.
\newblock


\bibitem[\protect\citeauthoryear{Shafahi, Huang, Najibi, Suciu, Studer,
  Dumitras, and Goldstein}{Shafahi et~al\mbox{.}}{2018}]%
        {shafahi2018poison}
\bibfield{author}{\bibinfo{person}{Ali Shafahi}, \bibinfo{person}{W~Ronny
  Huang}, \bibinfo{person}{Mahyar Najibi}, \bibinfo{person}{Octavian Suciu},
  \bibinfo{person}{Christoph Studer}, \bibinfo{person}{Tudor Dumitras}, {and}
  \bibinfo{person}{Tom Goldstein}.} \bibinfo{year}{2018}\natexlab{}.
\newblock \showarticletitle{Poison Frogs! Targeted Clean-label Poisoning
  Attacks on Neural Networks}. In \bibinfo{booktitle}{\emph{NeurIPS}}.
  \bibinfo{pages}{6106--6116}.
\newblock


\bibitem[\protect\citeauthoryear{Sheikh, Goel, and Kumar}{Sheikh
  et~al\mbox{.}}{2020}]%
        {sheikh2020approach}
\bibfield{author}{\bibinfo{person}{Mohammad~Ahmad Sheikh},
  \bibinfo{person}{Amit~Kumar Goel}, {and} \bibinfo{person}{Tapas Kumar}.}
  \bibinfo{year}{2020}\natexlab{}.
\newblock \showarticletitle{An approach for prediction of loan approval using
  machine learning algorithm}. In \bibinfo{booktitle}{\emph{2020 International
  Conference on Electronics and Sustainable Communication Systems}}.
  \bibinfo{pages}{490--494}.
\newblock


\bibitem[\protect\citeauthoryear{Shklovski, Mainwaring, Sk{\'u}lad{\'o}ttir,
  and Borgthorsson}{Shklovski et~al\mbox{.}}{2014}]%
        {shklovski2014leakiness}
\bibfield{author}{\bibinfo{person}{Irina Shklovski}, \bibinfo{person}{Scott~D
  Mainwaring}, \bibinfo{person}{Halla~Hrund Sk{\'u}lad{\'o}ttir}, {and}
  \bibinfo{person}{H{\"o}skuldur Borgthorsson}.}
  \bibinfo{year}{2014}\natexlab{}.
\newblock \showarticletitle{Leakiness and creepiness in app space: Perceptions
  of privacy and mobile app use}. In \bibinfo{booktitle}{\emph{SIGCHI}}.
  \bibinfo{pages}{2347--2356}.
\newblock


\bibitem[\protect\citeauthoryear{Shokri, Stronati, Song, and Shmatikov}{Shokri
  et~al\mbox{.}}{2017}]%
        {shokri2017membership}
\bibfield{author}{\bibinfo{person}{Reza Shokri}, \bibinfo{person}{Marco
  Stronati}, \bibinfo{person}{Congzheng Song}, {and} \bibinfo{person}{Vitaly
  Shmatikov}.} \bibinfo{year}{2017}\natexlab{}.
\newblock \showarticletitle{Membership Inference Attacks Against Machine
  Learning Models}. In \bibinfo{booktitle}{\emph{S\&P}}.
  \bibinfo{pages}{3--18}.
\newblock


\bibitem[\protect\citeauthoryear{Song, Ristenpart, and Shmatikov}{Song
  et~al\mbox{.}}{2017}]%
        {song2017machine}
\bibfield{author}{\bibinfo{person}{Congzheng Song}, \bibinfo{person}{Thomas
  Ristenpart}, {and} \bibinfo{person}{Vitaly Shmatikov}.}
  \bibinfo{year}{2017}\natexlab{}.
\newblock \showarticletitle{Machine Learning Models that Remember Too Much}. In
  \bibinfo{booktitle}{\emph{CCS}}. \bibinfo{pages}{587--601}.
\newblock


\bibitem[\protect\citeauthoryear{Szyller, Atli, Marchal, and Asokan}{Szyller
  et~al\mbox{.}}{2021}]%
        {szyller2021dawn}
\bibfield{author}{\bibinfo{person}{Sebastian Szyller},
  \bibinfo{person}{Buse~Gul Atli}, \bibinfo{person}{Samuel Marchal}, {and}
  \bibinfo{person}{N Asokan}.} \bibinfo{year}{2021}\natexlab{}.
\newblock \showarticletitle{Dawn: Dynamic adversarial watermarking of neural
  networks}. In \bibinfo{booktitle}{\emph{MM}}. \bibinfo{pages}{4417--4425}.
\newblock


\bibitem[\protect\citeauthoryear{Tipping and Bishop}{Tipping and
  Bishop}{1999}]%
        {tipping1999mixtures}
\bibfield{author}{\bibinfo{person}{Michael~E Tipping} {and}
  \bibinfo{person}{Christopher~M Bishop}.} \bibinfo{year}{1999}\natexlab{}.
\newblock \showarticletitle{Mixtures of Probabilistic Principal Component
  Analyzers}.
\newblock \bibinfo{journal}{\emph{Neural Computation}} \bibinfo{volume}{11},
  \bibinfo{number}{2} (\bibinfo{year}{1999}), \bibinfo{pages}{443--482}.
\newblock


\bibitem[\protect\citeauthoryear{Truex, Liu, Gursoy, Yu, and Wei}{Truex
  et~al\mbox{.}}{2021}]%
        {truex2021demystifying}
\bibfield{author}{\bibinfo{person}{Stacey Truex}, \bibinfo{person}{Ling Liu},
  \bibinfo{person}{Mehmet~Emre Gursoy}, \bibinfo{person}{Lei Yu}, {and}
  \bibinfo{person}{Wenqi Wei}.} \bibinfo{year}{2021}\natexlab{}.
\newblock \showarticletitle{Demystifying Membership Inference Attacks in
  Machine Learning As A Service}.
\newblock \bibinfo{journal}{\emph{TSC}} \bibinfo{volume}{14},
  \bibinfo{number}{06} (\bibinfo{year}{2021}), \bibinfo{pages}{2073--2089}.
\newblock


\bibitem[\protect\citeauthoryear{Uchida, Nagai, Sakazawa, and Satoh}{Uchida
  et~al\mbox{.}}{2017}]%
        {uchida2017embedding}
\bibfield{author}{\bibinfo{person}{Yusuke Uchida}, \bibinfo{person}{Yuki
  Nagai}, \bibinfo{person}{Shigeyuki Sakazawa}, {and}
  \bibinfo{person}{Shin'ichi Satoh}.} \bibinfo{year}{2017}\natexlab{}.
\newblock \showarticletitle{Embedding watermarks into deep neural networks}. In
  \bibinfo{booktitle}{\emph{ICMR}}. \bibinfo{pages}{269--277}.
\newblock


\bibitem[\protect\citeauthoryear{Wang, Wu, Zhang, and Yao}{Wang
  et~al\mbox{.}}{2020}]%
        {wang2020watermarking}
\bibfield{author}{\bibinfo{person}{Jiangfeng Wang}, \bibinfo{person}{Hanzhou
  Wu}, \bibinfo{person}{Xinpeng Zhang}, {and} \bibinfo{person}{Yuwei Yao}.}
  \bibinfo{year}{2020}\natexlab{}.
\newblock \showarticletitle{Watermarking in Deep Neural Networks via Error
  Back-Propagation}.
\newblock \bibinfo{journal}{\emph{Electronic Imaging}} \bibinfo{volume}{2020},
  \bibinfo{number}{4} (\bibinfo{year}{2020}), \bibinfo{pages}{22--1}.
\newblock


\bibitem[\protect\citeauthoryear{Wang and Wu}{Wang and Wu}{2022}]%
        {wang2022protecting}
\bibfield{author}{\bibinfo{person}{Yumin Wang} {and} \bibinfo{person}{Hanzhou
  Wu}.} \bibinfo{year}{2022}\natexlab{}.
\newblock \showarticletitle{Protecting the intellectual property of speaker
  recognition model by black-box watermarking in the frequency domain}.
\newblock \bibinfo{journal}{\emph{Symmetry}} \bibinfo{volume}{14},
  \bibinfo{number}{3} (\bibinfo{year}{2022}), \bibinfo{pages}{619}.
\newblock


\bibitem[\protect\citeauthoryear{Wu, Liu, Yao, and Zhang}{Wu
  et~al\mbox{.}}{2020}]%
        {wu2020watermarking}
\bibfield{author}{\bibinfo{person}{Hanzhou Wu}, \bibinfo{person}{Gen Liu},
  \bibinfo{person}{Yuwei Yao}, {and} \bibinfo{person}{Xinpeng Zhang}.}
  \bibinfo{year}{2020}\natexlab{}.
\newblock \showarticletitle{Watermarking neural networks with watermarked
  images}.
\newblock \bibinfo{journal}{\emph{TCSVT}} \bibinfo{volume}{31},
  \bibinfo{number}{7} (\bibinfo{year}{2020}), \bibinfo{pages}{2591--2601}.
\newblock


\bibitem[\protect\citeauthoryear{Xue, Zhang, Wang, and Liu}{Xue
  et~al\mbox{.}}{2021}]%
        {xue2021intellectual}
\bibfield{author}{\bibinfo{person}{Mingfu Xue}, \bibinfo{person}{Yushu Zhang},
  \bibinfo{person}{Jian Wang}, {and} \bibinfo{person}{Weiqiang Liu}.}
  \bibinfo{year}{2021}\natexlab{}.
\newblock \showarticletitle{Intellectual Property Protection for Deep Learning
  Models: Taxonomy, Methods, Attacks, and Evaluations}.
\newblock \bibinfo{journal}{\emph{TAI}} \bibinfo{volume}{3},
  \bibinfo{number}{6} (\bibinfo{year}{2021}), \bibinfo{pages}{908--923}.
\newblock


\bibitem[\protect\citeauthoryear{Yang, Lao, and Li}{Yang et~al\mbox{.}}{2021}]%
        {yang2021robust}
\bibfield{author}{\bibinfo{person}{Peng Yang}, \bibinfo{person}{Yingjie Lao},
  {and} \bibinfo{person}{Ping Li}.} \bibinfo{year}{2021}\natexlab{}.
\newblock \showarticletitle{Robust watermarking for deep neural networks via
  bi-level optimization}. In \bibinfo{booktitle}{\emph{ICCV}}.
  \bibinfo{pages}{14841--14850}.
\newblock


\bibitem[\protect\citeauthoryear{Zhang, Chen, Liao, Fang, Ma, Zhang, Hua, and
  Yu}{Zhang et~al\mbox{.}}{2021a}]%
        {zhang2021exploring}
\bibfield{author}{\bibinfo{person}{Jie Zhang}, \bibinfo{person}{Dongdong Chen},
  \bibinfo{person}{Jing Liao}, \bibinfo{person}{Han Fang},
  \bibinfo{person}{Zehua Ma}, \bibinfo{person}{Weiming Zhang},
  \bibinfo{person}{Gang Hua}, {and} \bibinfo{person}{Nenghai Yu}.}
  \bibinfo{year}{2021}\natexlab{a}.
\newblock \showarticletitle{Exploring structure consistency for deep model
  watermarking}.
\newblock \bibinfo{journal}{\emph{arXiv preprint arXiv:2108.02360}}
  (\bibinfo{year}{2021}).
\newblock


\bibitem[\protect\citeauthoryear{Zhang, Chen, Liao, Fang, Zhang, Zhou, Cui, and
  Yu}{Zhang et~al\mbox{.}}{2020}]%
        {zhang2020model}
\bibfield{author}{\bibinfo{person}{Jie Zhang}, \bibinfo{person}{Dongdong Chen},
  \bibinfo{person}{Jing Liao}, \bibinfo{person}{Han Fang},
  \bibinfo{person}{Weiming Zhang}, \bibinfo{person}{Wenbo Zhou},
  \bibinfo{person}{Hao Cui}, {and} \bibinfo{person}{Nenghai Yu}.}
  \bibinfo{year}{2020}\natexlab{}.
\newblock \showarticletitle{Model Watermarking for Image Processing Networks}.
  In \bibinfo{booktitle}{\emph{AAAI}}, Vol.~\bibinfo{volume}{34}.
  \bibinfo{pages}{12805--12812}.
\newblock


\bibitem[\protect\citeauthoryear{Zhang, Chen, Liao, Zhang, Feng, Hua, and
  Yu}{Zhang et~al\mbox{.}}{2021b}]%
        {zhang2021deep}
\bibfield{author}{\bibinfo{person}{Jie Zhang}, \bibinfo{person}{Dongdong Chen},
  \bibinfo{person}{Jing Liao}, \bibinfo{person}{Weiming Zhang},
  \bibinfo{person}{Huamin Feng}, \bibinfo{person}{Gang Hua}, {and}
  \bibinfo{person}{Nenghai Yu}.} \bibinfo{year}{2021}\natexlab{b}.
\newblock \showarticletitle{Deep Model Intellectual Property Protection via
  Deep Watermarking}.
\newblock \bibinfo{journal}{\emph{TPAMI}} \bibinfo{volume}{44},
  \bibinfo{number}{8} (\bibinfo{year}{2021}), \bibinfo{pages}{4005--4020}.
\newblock


\bibitem[\protect\citeauthoryear{Zhang, Gu, Jang, Wu, Stoecklin, Huang, and
  Molloy}{Zhang et~al\mbox{.}}{2018}]%
        {zhang2018protecting}
\bibfield{author}{\bibinfo{person}{Jialong Zhang}, \bibinfo{person}{Zhongshu
  Gu}, \bibinfo{person}{Jiyong Jang}, \bibinfo{person}{Hui Wu},
  \bibinfo{person}{Marc~Ph Stoecklin}, \bibinfo{person}{Heqing Huang}, {and}
  \bibinfo{person}{Ian Molloy}.} \bibinfo{year}{2018}\natexlab{}.
\newblock \showarticletitle{Protecting Intellectual Property of Deep Neural
  Networks with Watermarking}. In \bibinfo{booktitle}{\emph{AsiaCCS}}.
  \bibinfo{pages}{159--172}.
\newblock


\end{thebibliography}


\end{document}